\documentclass[a4paper,3p,12pt]{elsarticle}%review,
\usepackage[utf8]{inputenc}
\usepackage[T1]{fontenc}
\usepackage{cleveref}

\usepackage{graphicx}

\usepackage{amssymb}
\usepackage{dsfont}
\usepackage{url}
\usepackage[caption=false,font=normalsize,labelfont=sf,textfont=sf]{subfig}

\usepackage{xspace}	 % \xspace - automagically add space after macros
\newcommand{\var}[1]{\ensuremath{\textsf{#1}}\xspace} % oder \textnormal{}
\usepackage[ruled,noend,linesnumbered]{algorithm2e}

\usepackage{amsmath}

\journal{Computational Intelligence and Neuroscience}

\begin{document}

\begin{frontmatter}

\title{Self-Adaptation of Activity Recognition Systems to New Sensors}

\author[BS]{David~Bannach}
\author[UK]{Martin~J\"anicke}
\author[DFKI]{Vitor~F.~Rey}
\author[UK]{Sven Tomforde}
\author[UK]{Bernhard~Sick}
\author[DFKI]{Paul~Lukowicz}

\address[BS]{Bonsai Systems GmbH, c/o ETH Zürich, Institut für Elektronik, Zürich, Switzerland,\\
david@bonsai-systems.com}
\address[UK]{Intelligent Embedded Systems, University of Kassel, Kassel, Germany,\\
%+49 561 804 \{6020,6185\}, 
\{bsick,jaenicke,stomforde\}@uni-kassel.de}
\address[DFKI]{Embedded Intelligence, German Research Center for Artificial Intelligence, Kaiserslautern, Germany,\\
%+49 631 20575 4000, 
\{vitor.fortes,paul.lukowicz\}@dfki.de}

\begin{abstract}
	Traditional activity recognition systems work on the basis of training, taking a fixed set of sensors into account. In this article, we focus on the question how pattern recognition can leverage new information sources without any, or with minimal user input. Thus, we present an approach for opportunistic activity recognition, where ubiquitous sensors lead to dynamically changing input spaces. Our method is a variation of well-established principles of machine learning, relying on unsupervised clustering to discover structure in data and inferring cluster labels from a small number of labeled dates in a semi-supervised manner. Elaborating the challenges, evaluations of over 3000 sensor combinations from three multi-user experiments are presented in detail and show the potential benefit of our approach.
\end{abstract}


\end{frontmatter}

% \linenumbers

% !TEX encoding = UTF-8 Unicode

\section{Introduction}\label{sec:intro}

Today, state-of-the-art approaches to activity and context recognition typically assume fixed, narrowly defined system configurations dedicated to often also narrowly defined tasks.
Such systems can only work when sensors are known in the training phase and they cannot adapt to new sensors in their environment.
In turn, sensors are evermore present in our life, although not always available. When moving around, a person may face highly instrumented environments and places with little or no intelligent infrastructure. 
Concerning on-body sensing, a user may carry a varying collection of sensor enabled devices (mobile phone, watch, headset, etc.) on different, dynamically varying body locations
(different pockets, wrist, bag). Thus, in order to realize their full potential, systems need
to take advantage of devices that just ``happen'' to be in the environment, taking into account
their current placement and relevance.

%Today, given the high cost of implementing and training activity recognition systems, their use
%is very limited, since its costs outweigh the benefits[]. Thus, this new field of opportunistic activity
%recognition[] has the potential to (...).

In our previous work, we investigated how on-body position and orientation of on-body sensors can be inferred \cite{onbody,orientation}, how position shifts can be tolerated \cite{kaiposition}, and how one sensor can replace another \cite{magnetic}. In \cite{jaenicke14self}, we started to work on the challenge of seamless sensor integration. More precisely, this means to answer the question how can a new sensor's data be integrated in an existing activity recognition system at runtime in order to improve this recognition process. Extending a system that used $n$ sensors to one that uses $(n+1)$ has many challenges. For instance, training data is expensive, and thus we cannot expect the new $(n+1)$ data to be labeled. Moreover, the same activity can be performed in multiple different ways, i.e., training data may vary significantly from one day to the other.  

In this article, we tackle some of these problems, presenting a method that can include new sensors to an existing system in an unsupervised manner. This is achieved by using concepts and techniques from the field of unsupervised learning. In a nutshell, we improve the current $n$-dimensional system by assuming that the structure in the new $n+1$ dimension is related to the class distribution. Of course, this assumption may not always hold, leading to a possible downgrade in system performance. In order to counter this issue, we evaluate the balance between possible gain and conformance to the class distribution in the $n$ sensor data. We also deal with the variance in training data by applying standard machine learning techniques such as bagging. While improvement is not guaranteed, we show that in practice our method can do more good than harm, providing an overall improvement even when labeling data is scarce. Furthermore, we study the performance in the presence of user input, which can be used to validate the system.

This article presents a first, important step towards a method for self-adapting multi-sensor systems. In our experiments, we extend a one-dimensional to a two-dimensional system, but do not yet answer the question when to actually integrate a new sensor once it became available. These questions are in the focus of our ongoing research in this field.
 
The remainder of this article is organized as follows: Section \ref{sec:related_work} analyzes related work, before Section \ref{sec:challenges} sketches the key challenges and solution ideas. Our evaluation strategy is set out in Section \ref{sec:evaluation}.  The basic self-adaptation method is explained in Section \ref{sec:label_inferring}, while Sections \ref{sec:similarity_search} to \ref{sec:fault_reduction} focus on four concerted extensions of this basic method. Section \ref{sec:conclusion} summarizes the key findings and gives an outlook to our future research.

%This paper is divided thus. In (), we review the state of the art in (), followed by a description
%of our method in Section ().  In Section () we present our evaluation strategy, while in section ()
%we present and discuss the results obtained both in synthetic and real life datasets. Finally, Section
%() contains our conclusions and future work.

\section{Related Work}
\label{sec:related_work}
%The following paragraph gives an overview of the fields our applied methods originate from.
%After naming a suitable field of research, different aspects that are incorporated in our approach to realize a self-adaptive system design are discussed.

% field of research
The incorporation of new information sources can be interpreted as a form of {\em transductive transfer learning} as described in~\cite{Pan:2010dm}.
This field covers the general problem of transferring knowledge learned in one domain to another, eventually even with a different task.
Within the authors description in~\cite{Pan:2010dm}, a domain comprises a feature space and a marginal probability distribution for the occurrence of samples therein, as well as a task, i.e.\ classification, composed of a label space and an objective predictive function (the classifier).

Due to the dimensionality of the input spaces in our sensor adaptation, the domains are different, whereas the classification task stays the same, even with the utilization of new information sources. Even though the authors give a more restrictive definition on {\em domain adaptation} than usual (cf., e.g., \cite{jiang2008literature, daume2009frust}), our approach can be considered as a special case of this area.

Daumé also presented an approach for domain adaptation which takes labeled samples from both domains to construct an augmented feature space and uses the result as input to a standard learning algorithm~\cite{DaumeIII:2007wo}. Unlike our approach, the article requires both domains to have the same features and dimensionality. 

Duan et al.~\cite{Duan:2012wg} proposed a domain adaptation method for heterogeneous feature spaces. They first project the source and target feature spaces into a common subspace on which they apply a standard learning algorithm. The authors achieved promising results on a computer vision data set with 800 and 600 dimensional feature spaces. In contrast, our work explicitly assumes the addition of additional sensors, i.e., the target space contains additional features and the common subspace is equal to the source feature space. Furthermore, the feature spaces in activity recognition scenarios usually have significantly less dimensions, as each dimension is related to a signal from a physical sensor and the features (derived from these signals) are often selected manually.

% data
The need for a large amount of annotated training data has widely been recognized as a major issue for the practical deployment of activity recognition systems. As a consequence, different approaches were studied to overcome this problem.
One line of work looks at unsupervised discovery of structure in sensor data (e.g., \cite{minnen2007discovering, huynh2006unsupervised}), while others attempt to develop activity models from online information, such as a common sense data base, WordNet, or general ``how to'' resources  (e.g., \cite{wyatt2005unsupervised, tapia2006building, chen2009ontology, stikic2008exploring}).
%Due to the nature of the data, such work has a strong focus on interaction with objects.

Beyond fully unsupervised approaches there has also been considerable interest in semi-supervised systems that limit the amount of data needed for training (e.g., \cite{stikic2008exploring,guan2007activity}). In general, the fields of \textit{semi-supervised learning} (SSL) and -- to a lower extent -- also \textit{active learning} (AL) \cite{CalmaRSL15} are relevant to the work presented in this article.

SSL~\cite{RV95,chapelle2006semi} makes use of both labeled and unlabeled data to train a prediction model. Therefore, SSL falls between unsupervised learning (without any labeled training data) and supervised learning (with completely labeled training data).
SSL may be based on generative models such as probabilistic models (e.g., Gaussian mixture models), low density separation algorithms, or graph-based methods, for instance.
The best known example for the second are transductive support vector machines (TSVM)~\cite{BD98} that build a connection between a density model for the data and the discriminative decision boundary by putting the boundary in sparse regions.
Doing so, TSVM uses unlabeled data in a semi-supervised manner.
A typical example for the third class are Laplacian SVM~\cite{GCMC08} that use graph based models, so called Lapalacian graphs, for semi-supervised learning.
A detailed overview of semi-supervised learning is given in~\cite{zhu2006semi}.

In the field of AL, membership query learning (MQL)~\cite{Angluin88}, stream-based active learning (SAL)~\cite{ACLEM90}, and pool-based active learning (PAL)~\cite{LG94} are the most important learning paradigms.
MQL is not relevant here because it may generate artificial samples that cannot be understood or labeled by human experts~\cite{Settles09}.
Similar to our work, SAL focuses on sample streams. That is, each sample is typically ``drawn'' once, and the system must decide whether to query the expert or to discard the sample~\cite{Settles09}.
PAL builds a ranking on a given pool of unlabeled samples and, depending on a certain selection strategy, chooses a set of samples that must be labeled.
A number of different selection strategies exist for that purpose, and four main categories can be distinguished: uncertainty sampling strategies select samples for which the considered classifier is most uncertain~\cite{TK02, CL06, SC08}, density weighting strategies consider the samples' distribution in the input space~\cite{NS04, DCB07, SC08}, estimated error reduction strategies aim at reducing the generalization error of the classifier directly~\cite{MMP04, GG07}, and diversity sampling strategies prevent the selection of ``redundant'' samples if more than one sample is selected in each query round~\cite{DRH06}.

In~\cite{Dai:2007tu}, Dai et al.\ applied co-clustering to train a classifier for unlabeled documents of a target domain when only labels for documents in a (different) source domain are known. Co-clustering can reveal features that are similar and may help classifying items in different domains, yet it is not capable of using new features to improve classification. Co-training~\cite{Blum:1998ua} is not applicable in our case because source and target feature spaces are not conditionally independent (the target space contains all features of the source space) and the goal is not to extend a training data set, but to expand the classifier to an extended input space.

Closest to our work is \cite{calatroni2010methodology}, which uses other sensors and behavioral assumptions to train a system to take advantage of a new sensor.
However, unlike our work, it does not consider adding a new sensor to an existing system. 
 
While related to our work at a methodological level, none of the articles mentioned above addresses the problem of integrating new sensors into an existing system. Consequently, our work aims at a continuous self-improvement process as part of activity recognition systems. In this context, it reflects the motivation of systems with increasingly dynamic and heterogeneous system compositions \cite{THSRSWS14,THS14} which is within the scope of research initiatives dealing with self-management and autonomous self-adaptation as described, e.g., in \cite{MSSU11}.

\section{Key  Challenges and Ideas}
\label{sec:challenges}

In order to be able to deal with adaptive sensor integration in terms of self-adaptation of activity recognition systems, we propose to combine efforts from the domain of semi-supervised learning:
\begin{enumerate}
  \item Unsupervised clustering to discover structure in data.
  \item We assume that structure corresponds to class membership. In the simplest case, this means that points in the same cluster belong to the same class. 
  \item We need to obtain at least one labeled data point for each cluster to label that cluster.
\end{enumerate}
Our approach uses a classifier trained on $N$ features to provide labeled data points for a ($N{+}1$) dimensional feature space. For the sake of simplicity and better understanding, we consider the transition from an one-dimensional (1D -- a single sensor) to a two-dimensional (2D -- two sensors) space here.

We start with a classifier trained on data from a single sensor (sensor~1). We assume that the classifier is performing reasonably well, but still has some regions in its feature space where the separation of classes is poor. For our method it is crucial that it is known in which regions of the feature space the classifier is confident and in which it has many misclassifications. In a tree classifier (which we are using for this work), this is trivially given if the purity of the leaves is retained from training. 

At a certain stage, a second sensor appears (sensor~2), and this sensor can potentially improve the overall class separation. We can now collect and cluster data points in the 2D space, but a remaining question is how to assign labels to the 2D clusters. 
The aim of our work is to do so without or with minimal user input. Thus, we use the classifier trained on the 1D feature space to provide the labels. A naive method to do this proceeds as follows (see Fig.\ \ref{fig:inferring}):
\begin{enumerate}
\item Project the 2D clusters onto the 1D space of sensor~1.
\item Consider regions of the 1D space onto which points from {\em only a single 2D cluster} are projected. From the assumption about structure corresponding to class distribution, we can conclude that the class distribution within those regions should be representative for the class distribution in the corresponding clusters.
\item Use the classifier trained on sensor~1 to determine the class distribution in the single cluster regions and use this distribution to label the corresponding clusters.   
\end{enumerate}
As can be seen in Fig.\ \ref{fig:inferring}, the performance gain comes from clusters that are partially projected onto regions where the 1D classifier is confident and partially onto regions where it produces many errors. 
For the former case, the 1D classifier provides labels for the 2D clusters.
For the latter case, the 2D classifier outperforms the 1D classifier. 

Note that the basic method does not necessarily require a one-to-one correspondence between clusters and classes.
As an example, consider a region that in a 1D space corresponds to a 50:50 mixture of two classes. Now assume that in 2D that region corresponds to two clusters: one with a 25:75,  and one with a 75:25 mixture of those two classes. It then follows that using the cluster information reduces the
probability of misclassification from  50\% to just 25\% in that region.

\newcommand{\figheight}{8.5cm}
\begin{figure}[htb]
\begin{center}
%\hspace*{-2mm}
\subfloat[label inferring]{
  \includegraphics[height=\figheight]{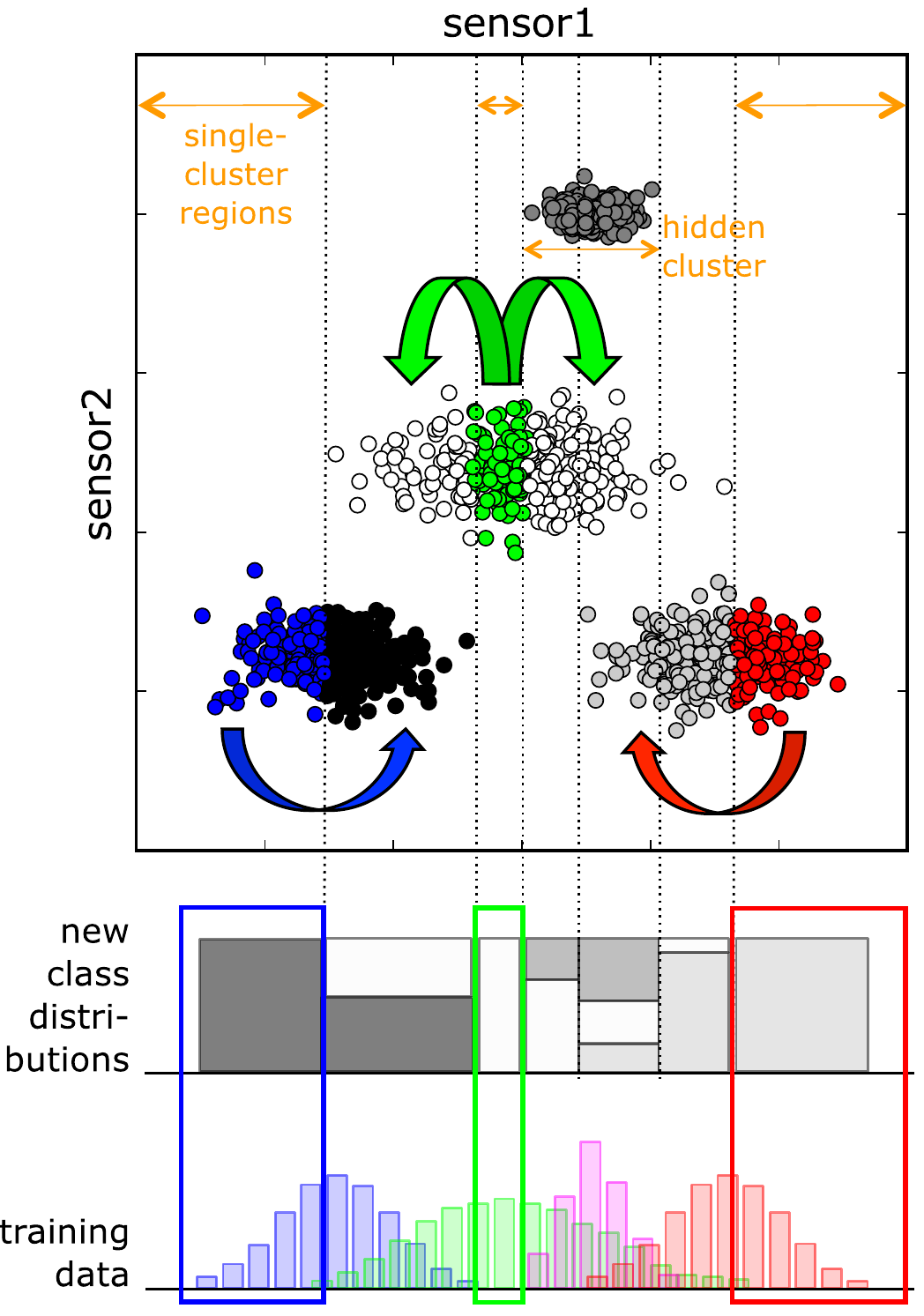}
  \label{fig:inferring}
  \label{fig:clusters}
}
%\hspace*{-3mm}
\subfloat[similarity search]{
  \includegraphics[height=\figheight]{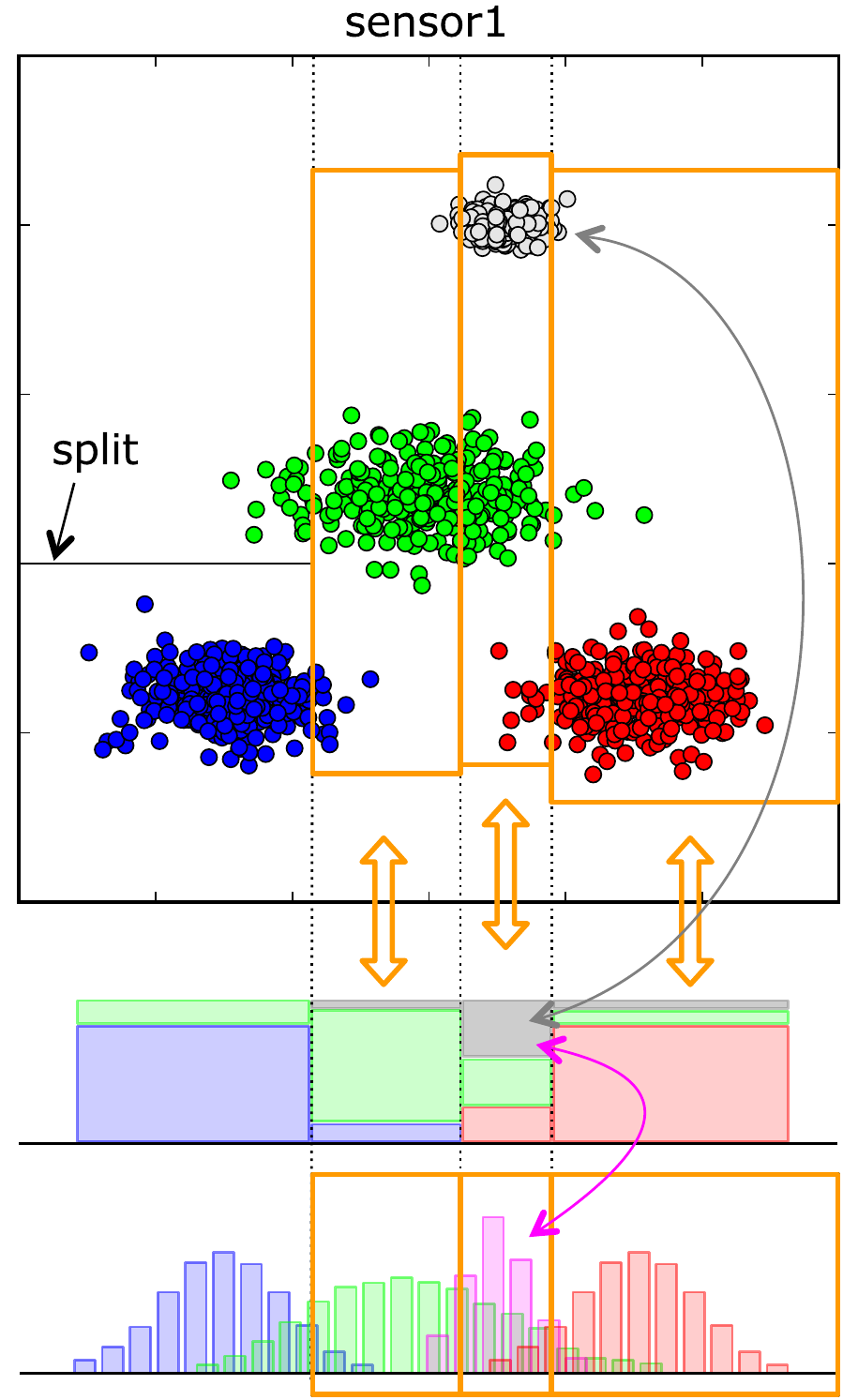}
  \label{fig:similaritysearch}
}
\caption{Illustration of methods for assigning labels from 1D feature space to 2D clusters.
	Dashed lines: boundaries of tree trained according to (a) cluster membership 
	and (b) training data.
}
\label{fig:clusterlabeling}
\end{center}
\end{figure}

\subsection{Challenges}

Clearly, the method works well on some of the sample clusters in Fig.\ \ref{fig:inferring} -- because the 1D and 2D distributions were carefully constructed. In most real data sets, such ideal distributions are unlikely. When applied to these data, the naive method described above is not able to achieve any improvement because the clusters that it can label using the ``pure'' regions of the classifier do not appear together with any ``impure'' region that could be improved. Furthermore, the variation of distributions between two samples (unlabeled 2D data, 2D test data) can be significant, which may even lead to a performance decrease.

In the remainder of this section, we will systematically outline the challenges to be addressed by a more sophisticated method.

\subsubsection{Distribution Issues}
\label{sec:distribution}

The basic method requires that each cluster has a region where, when projected onto the 1D space, it does not overlap with any other cluster. Unfortunately, this does not always hold. Some clusters can be hidden, as shown in Fig.\ \ref{fig:similaritysearch}. To assign labels to such hidden clusters, we propose a ``similarity search''. To this end we, first  assign labels to all non hidden clusters. 

We then attempt to assign labels to the hidden clusters in such a way that, when the labels are propagated to the cluster's data points, the projection of all data points onto 1D space approximates the original 1D distribution (as trained for the 1D classifier) as good as possible.

Another distribution-related issue addresses the specific choice of a tree classifier, which stores class distributions at leaf level. This means that in order to be able to assign a class distribution to a cluster, that cluster must be the only cluster appearing in the leaf. Unfortunately, in real data we often have the situation where two or more clusters are projected onto distinct areas of the 1D space, but these areas fall within the same leaf making class distribution assignment impossible.
% solution to distribution related issue
To address this problem, we create an additional tree in the 1D feature space using purity with respect to cluster membership rather than class membership as splitting criterion.
We then use the original training data for the 1D classifier to assign class distributions to the new leaves which are in turn used to assign distributions to the clusters.

\subsubsection{Training Sample Quality and Size}

All machine learning algorithms build on the assumption that the training data is reasonably representative and break down if that assumption is substantially violated. In activity recognition, training data variance is often a problem because (1) it is costly to obtain a large amount of labeled data from users and (2) there are often large variations in the  way humans perform even fairly simple actions. At the same time, our method is particularly sensitive to the quality of the training data, because it leverages small areas of the feature space (regions of the 1D space onto which only parts of a single cluster are projected) to label larger ones (entire clusters). Using small areas to derive labels for larger ones effectively means artificially reducing the sample size and thus making the estimator more sensitive to sample variance. In practice, we would often use 10\% and less of the data for cluster labeling. If that 10\% of the training data happens to have high variance and if they are not representative, % of the later test input 
then the addition of the new sensor will not only fail to improve performance, it may actually drastically reduce the classification accuracy over the entire feature space.   
As described later, we have found such ``erroneous'' configurations to be a significant problem in real-world activity recognition data. In fact, if no additional measures are taken, the risk of ending up with  erroneous configurations that worsen the recognition performance nearly neutralizes the potential improvement from the integration of an additional sensor. To solve this problem, we have developed three approaches for discarding configurations that have a high probability of leading to performance degradation. 
First, for every assignment of class distributions to clusters, we compute \emph{plausibility} and \emph{gain} values. The plausibility is based on the projections of the labeled clusters onto the 1D feature space. It compares the class distribution that results from the projection to the class distribution in the original 1D training data, and thus offers a measure of how plausible the guessed labels are. The gain compares the ``purity'' of the labeled clusters to the ``purity'' of the corresponding leaves of the original classifier.
It is an estimation of the potential that the particular cluster labeling has for improving the system performance. The larger the discrepancy between the two measures, the more likely it is that the specific cluster labeling has been caused by sample variance and does not correspond to the true class distribution. 

Overall, gain and plausibility proved very useful in deciding whether it makes sense to use a particular cluster labeling or not. However,
%they are just heuristics and 
it is well possible that a solution with high plausibility and high gain is distorted by sample variance and will worsen rather than improve performance (as compared to the 1D classifier). As an additional measure to address this problem, we rely on bagging. That is, we derive the cluster structure and its class distribution for different subsets of the training data created from the original set through randomly leaving out and replicating individual samples \cite{bagging}. We consider clusters and class distribution assignments that significantly vary between the subsets to be unstable with respect to training data variance and penalize their plausibility value.  

As a final measure, we consider user-provided additional labels to detect erroneous configurations. We apply both, the original 1D classifier and the 2D classifier generated by our method, to new data points and compare the output. If the output differs, we ask the user to provide a label to check if the new classifier would improve (if it was right) or worsen (if it was wrong) the performance (cf.\ selection strategies for stream-based active learning). We reject the new classifier as erroneous if it makes more than a pre-defined number of mistakes within a certain number of data points. When looking at points where the classifiers differ, even few user-provided labels have a very high probability of spotting erroneous configurations with little (but non-zero) probability of mistakenly rejecting those that improve the classification rate by a relevant factor.

\subsubsection{Clustering Issues}
In Fig.\ \ref{fig:clusterlabeling}, the optimal clustering is obvious. However, even in such a simple case standard clustering algorithms would only return the ``optimal'' structure if they were provided with the correct number of clusters and managed to avoid suboptimal configurations related to local minima. In general neither can be assumed. 
Thus, we rely on a hierarchical clustering algorithm to provide a set of possible configurations, but choose the most promising one by generating cluster labels for each configuration and evaluating their plausibility and gain metrics.

\subsection{Method Overview}

We base the work described in this article on a tree classifier. In a tree classifier, each region of the classifier is represented by a single leaf of the classifier tree, and its ancestor nodes define the region's boundaries (hyperplanes, each perpendicular to one axis). Additional dimensions can easily be added by simply replacing a leaf with a new subtree which adds boundaries on the extra axis without affecting other regions.

\begin{figure}[htbp]
\begin{center}
\includegraphics[width=\textwidth]{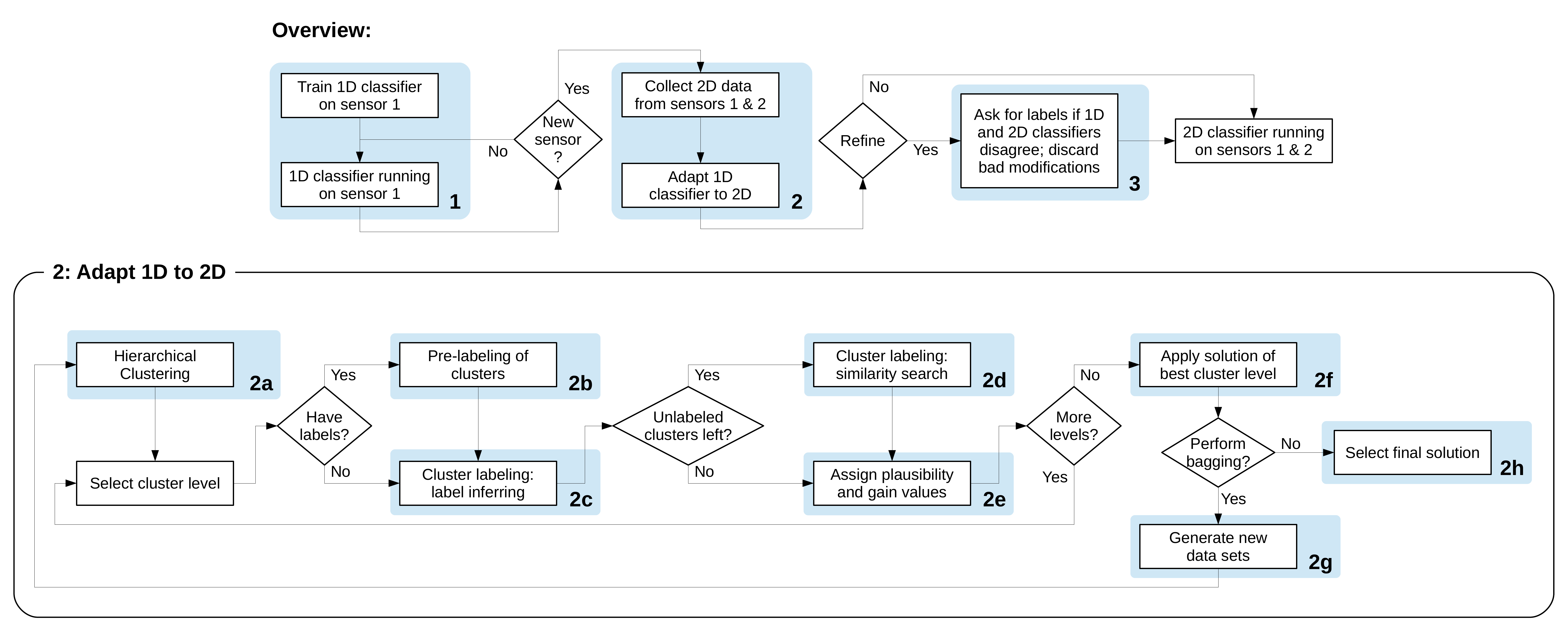}
\caption{Overview of the overall procedure (top) and the most important step, step 2, the classifier adaptation (bottom).}
\label{fig:flowchart}
\end{center}
\end{figure}

The description above already indicates that our proposed method consists of three stages, see Fig.\ \ref{fig:flowchart} (top):
\begin{enumerate}
\item Training and using an 1D tree classifier on sensor~1 in a standard supervised manner.
\item Collecting and clustering 2D data from sensor~1 and sensor~2 and using the classifier trained on sensor~1 to label the 2D clusters.
\item Optionally running the 1D and 2D classifiers in parallel on additional data points and collecting labels whenever they disagree to sort out potentially erroneous configurations (Section~\ref{sec:fault_reduction}).
\end{enumerate}
After processing the previous steps, the 2D classifier is ready for use. The core part (point two) consists of the following sub-stages, see Fig.\ \ref{fig:flowchart} (bottom):
\begin{enumerate}
\item Running a hierarchical clustering algorithm on the 2D data.
\item For each plausible cluster configuration:
\begin{enumerate}
\item If desired using any (usually very few) labels provided by the user for some 2D data points to improve the labeling (Section~\ref{sec:prelabeling}).
\item Label inferring: assigning class distributions to clusters using those areas of the 1D feature space onto which only a single cluster is projected (Section~\ref{sec:label_inferring}). This involves creating a new 1D tree trained according to cluster rather than class membership (Section~\ref{sec:inferring_cluster_labels}). 
\item Using similarity search to label hidden clusters where possible (Section~\ref{sec:similarity_search}). 
\item Assigning plausibility and gain values for the resulting class distribution assignment for each region of the 1D classifier (Section~\ref{sec:plausibility_and_gain}).
\end{enumerate}
\item Selecting the clustering and label assignment that have the plausibility and gain measures best matching a predefined policy.
\item If required, generating a new data set using the bagging method and going back to Step~1 (Section~\ref{sec:bagging}).
\item Selecting the final 2D classifier configuration.
\end{enumerate}
Once the final 2D classifier is ready, we evaluate it as described in the next section.

\section{Evaluation Strategy on Real Data Sets}
\label{sec:evaluation}

We evaluate our method on data from previously published data sets: a bicycle repair data set \cite{bicycle1}, a car maintenance data set \cite{skoda}, and the OPPORTUNITY data set \cite{opportunity}. Over the three data sets, we study 230 different sensor combinations from 13 users executing 22 different activities (8 in the OPPORTUNITY, 9 in the bicycle, and 5 in the car maintenance data set) 
% (2*132 + 3*56 + 8*42)*5 sensor combinations
% 345-32 + 332-32 + 515-54 + 510-54 + 506-54 + 411-277 + 5*(427-277) + 426-277 + 397-277 = 3135
with over 3000 individual activity instances. The sensors in question have widely varying recognition rates: from 21.89\% to 94.15\% with a mean of 51.05\% for an individual sensor and from 20.26\% to 97.66\% with a mean of 69.25\% for a combination of two sensors trained in a fully supervised manner as baseline. 
  
While the bicycle task has fairly complex, longer actions (e.g., taking out the seat, disassembling the wheel), the car maintenance and OPPORTUNITY data contains mostly different types of very short open/close gestures. 

As features on the sensors described below we allow only mean and variance. Due to this setting, we decouple the problem investigated in this work from the issue of feature choice which can be complex and has a big influence on the recognition success. At the same time, both features are commonly used in motion-related activity recognition. Given an unknown sensor, the system would certainly not be able to identify sophisticated features and would have to work with what is known to work reasonably well across many sensors.

We evaluate our method on each combination of two different sensors; we train the base recognition system on the first and then integrate the second one as the ``new sensor''.
%
%For the sake of simplicity 
As mentioned earlier, we treat each feature as a single sensor in this evaluation. Thus, we also consider improving a classifier with another feature of the same sensor or even with the same feature on a different axis of the sensor. Furthermore, this includes cross-modality situations where a system trained on one sensor (e.g., accelerometer) is improved with a sensor of a completely different modality (e.g., radio frequency positioning). 

%best A
In order to avoid improving only the shortcomings of a badly trained recognition system, we chose the parameters for the base classifier (minimum number of instances per leaf) individually for each combination such that the base system and the comparison system trained on both features perform best on the training set.

%\subsection{Evaluation Methodology}
%\label{sec:methodology}

For evaluation purposes, we select the C4.5 trained decision tree classifier (WEKA J48 implementation, binary decisions) as base classifier. We normalize features from real data sets to the unit interval $[0,1]$. For each evaluation, we need three separate data sets: for \emph{training}, \emph{adapting}, and \emph{testing}. The training data set is needed for training the initial classifier and uses only one feature and labels. The data set that we need for adapting the original classifier has two features but no labels, and the data set we use for testing the performance has both features plus the labels. If not stated otherwise, we set the size of the adapting and test data sets both to half the size of the training data set. For each sensor combination, we evaluate five different splittings by shuffling the data each time before splitting into the three sets.

\section{Label Inference}
%\section{Inferring Labels from Initial Classifier Model}
\label{sec:label_inferring}

As a first step, we focus on the inference of cluster labels from single cluster regions within the projection of 2D cluster onto the 1D space of the original sensor. In Fig.\ \ref{fig:flowchart} this corresponds to steps 1 (clustering), 2b (cluster label inferring), 2d (plausibility and gain computation), and 3 (taking the solution with best gain/plausibility ratio).

\subsection{Finding Structure in Extended Feature Space}
\label{sec:clustering}
In order to assess the structure of the extended feature space, we apply an agglomerative, hierarchical clustering algorithm. The resulting cluster tree gives us the possibility to explore the structure at different levels of detail without deciding on the number of clusters yet. We traverse the cluster tree top down, starting with two clusters and splitting one cluster at each step until a maximum number of clusters is reached. At each step, we attempt to (1) infer a label distribution for each cluster based on information from the original classifier model, and (2) build an adapted classifier model on the extended feature space, utilizing the label information of the clusters. All those adapted models are then rated based on the possible gain of classification accuracy and the best one is selected.

The method of rating different solutions not only allows for automatically finding an appropriate number of clusters but also makes it possible to choose the most helpful among different features available for a sensor, or even among different sensors.

\subsection{Inferring Cluster Labels}
\label{sec:inferring_cluster_labels}
\label{sec:multicluster}

As described in Section \ref{sec:challenges}, the basic method for assigning labels to clusters is based on projecting the clustered data points back onto the original feature space. We then identify regions of the original feature space where (1) a single cluster is projected onto and (2) where multiple clusters are projected onto. We refer to them as \emph{single-cluster} and \emph{multi-cluster} regions (see Fig.\ \ref{fig:clusters}). In single-cluster regions, the class distribution found in the original model (or in the training data) for that region is directly related to the cluster and may be used to infer the cluster label distribution. In multi-cluster regions, however, it is not clear how the class distribution in that region is correctly partitioned among the involved clusters, unless the distribution is pure, i.e., contains a single class only.

The challenges that have to be addressed are (1) delimiting the single- and multi-cluster regions and (2) transferring the class distribution from the original classifier onto the single-cluster regions. 

The first challenge, distinguishing between single- and multi-cluster regions, is in essence a classification task by itself. We found the C4.5 decision tree classifier to be a good choice for solving this problem. The amount of pruning ($B=$ minimal number of data points in a region/leaf) controls the granularity and stability against noise. Thus, for the actual process of cluster labeling, we project the clustered data points onto the original feature space and train a C4.5 decision tree based on cluster membership. All regions (leaves) in this tree with a purity above a preset \emph{purity threshold} (0.95 proved to be a good choice) are treated as single-cluster regions.

For label assignment, we then have to go back to the training data used for the original classifier. We use it to estimate the class density in those leaves of the tree trained on the clusters that correspond to single-cluster regions. In a final step, the label distribution is transferred to the corresponding clusters and normalized.

%Note that the above means that we need to retain either the original
%training data or at least an adequate representation of it to be able
%to extend the system with a new sensor using our method.  

\subsection{Adapting the Classifier Model to the Extended Feature Space}
\label{sec:adaptation}

With clusters that appear in single-cluster regions at hand and label distributions assigned to them, these distributions can be used to extend the classification model with the new dimension. Obviously, an extension only makes sense for the multi-cluster regions of the original feature space, since in the single-cluster regions the class distribution is identical in the new and in the old feature spaces. Furthermore, the extension is only possible for those multi-cluster regions in which all clusters have been labeled. As explained in Section \ref{sec:distribution} and Fig.\ \ref{fig:clusters}, some clusters may be hidden with no points in the cluster being projected onto a single cluster region (which is the prerequisite for the ability to assign a class distribution to a cluster). 

Thus, multi-cluster regions in which all clusters have been labeled are split according to the cluster membership of the points. For a decision tree classifier, this means that each leaf of the original classifier which falls into such a multi-cluster region is replaced by a subtree with each leaf of the subtree corresponding to a different cluster.

\subsection{Estimating Plausibility and Gain}
\label{sec:plausibility_and_gain}
In essence, the above method amounts to generalizing the class distribution from a small subset of points (the points that project onto single-cluster regions) to the entire cluster from which the subset has been taken. For such a generalization becoming valid, two conditions must be fulfilled.

First, the class distribution must be homogeneous throughout the cluster, i.e., when split into separate regions, the class distribution must be the same in each region. Note that it is not necessary that all points within the cluster belong to the same class (although it is certainly desirable). Instead, the probability of a point belonging to a given class must be (approximately) the same everywhere within the cluster. The main questions are (a) ``Does the new feature space contain enough correspondence between structure and class distribution to find clusters with homogeneous class distribution?'' and (b) ``Is our clustering algorithm able to find those clusters?''. 

Second, the subset of points that we use for labeling must be a good representation of the class distribution within the cluster. The main concerns are the sample size in relation to sample variance and the actual homogeneity of the distribution within the cluster. We may thus envisage a situation where, overall, the class distribution within a cluster is homogeneous enough to justify assigning a single distribution to the cluster, but between the small single-cluster regions, from which we collect the labels, the variations would be considerable. Clearly, such regions should not be used for the derivation of the class distribution. 

If the sensor used to create the new feature space contains additional information about the classification problem at hand, it is well possible that the above requirements are (at least to a certain degree) fulfilled. However, this is by no means guaranteed. Thus, for example, there is no fundamental reason why an additional sensor should not result in a distribution that has two spatially well separated clusters, but where the separation boundary between classes is located within the clusters instead of between them. Furthermore, even if in the new feature space there exists a cluster structure that satisfies the above conditions, there is no way to guarantee that our clustering algorithm will find it.

Unfortunately, applying our method to a distribution that does not satisfy the above conditions in general leads to a significant performance decrease (i.e., worse than just a lack of improvement). At the same time, the proposed system should be able to deal with an arbitrary additional sensor without prior knowledge about the class distribution in the new feature space. The problem herewith is that, for a given clustering and label distribution in the new feature space, there is no way to {\em exactly} determine (without additional user provided labels) the quality of the label assignment. As a consequence, our system relies on the following heuristic:
\begin{enumerate}
\item A method for the estimation of the \emph{plausibility} of label assignment.
\item A method for the estimation of the potential \emph{gain} of accuracy that the new dimension may bring.
\item A risk-benefit analysis to decide a given label assignment is worth using.
\end{enumerate}

\subsubsection{Plausibility}
\label{sec:plausibility}
The derivation of  \emph{plausibility} is based on the observation, that---while the inclusion of the new sensor changes the classification of individual data points---the overall class distribution within each region of the old feature space should remain unchanged. Thus, we propose to take a set of data points from the extended feature space and
\begin{enumerate}
\item classify them using the new (extended feature space) decision tree,
\item project them onto the old feature space, and
\item for each region in the old feature space:
\begin{enumerate}
  \item compute the resulting class distribution in the old feature space, and 
  \item compare it to the class distribution in the  original training data on the old feature space.
\end{enumerate} 
\end{enumerate} 
If the labels assigned to the clusters in the new feature space are correct, then the two distributions (point 3a and point 3b above) should be identical or at least very similar. Therefore, we derive the plausibility value for each region $r$ from this similarity as
%\begin{equation}
\[ \var{plausibility}_{r} = \frac{1}{2}\sum\limits_{i=0}^N (\hat{p}_{r,i}-p_{r,i})^{2} , \]
%\end{equation}
%\newcomment{BS@DB: Hier wuerde man ein Divergenzmass wie Hellinger oder Kullback-Leibler erwarten...}
where $p_{r,i}$ is the probability of class $i$ (estimated using the training data) within region $r$ and $\hat{p}_{r,i}$ the probability of class $i$ in region $r$ approximated with data in the extended feature space when classified with the adapted classifier. The plausibility is normalized by the maximal possible distance of the two class distributions.

Note that the plausibility value is just a heuristic. If the value is high, then (within sample variations) we can be sure that the label assignment in the new data space is not useful. On the other hand, when it is low it is likely (but not necessarily assured) that label assignment is approximately correct.

\subsubsection{Gain}
\label{sec:gain}
As a second measure we estimate the potential classification accuracy \emph{gain}.   
Since we consider a tree classifier, this can be defined as an increase of purity that we achieve by extending the classifier with the new sensor. As described in \Cref{sec:adaptation}, we create the extended classifier by splitting leaves of the old classification tree according to membership in (labeled) clusters of the new feature space.  For each leaf of the old tree (i.e., for each region $r$ in the initial feature space), we can compute its purity from the training data as 
\[ \var{purity}_{r} = \max_{i}(p_{r,i}). \]
In the best case, the new leaves in a region are all pure and, hence, the new purity would be~$1$. However, if a new leaf would still classify some points incorrectly (which we cannot test without ground truth data) the relative size of the leaf (in terms of amount of data points falling in that leaf) would be larger than the probability of the corresponding class in the original training data within that region.
Note that multiple new leaves are considered together if they contribute to the same class. Hence, the purity contribution of a new leaf can maximally amount to the probability of its corresponding class in the original training data. Thus, we obtain the purity resulting from the new leaves in each region from the histogram intersection of the new class distribution (point 3a above) with the initial class distribution (point 3b above) 
\[ \widehat{\var{purity}}_{r} = \sum\limits_{i=0}^N \min(\hat{p}_{r,i},p_{r,i}), \]
and we finally obtain the gain value from the difference of both purities
\[ \var{gain}_{r} = \widehat{\var{purity}}_{r} - \var{purity}_{r}. \]

The gain value is computed separately for each region of the original classifier and is summed up to a weighted average for the whole solution. The weights are the relative amount of training data points in each region.

Note that the gain computation is based on the assumption that the label assignments in the new feature space are correct. In this sense, it is (just like plausibility) an estimation that may or may not be correct.

\subsubsection{Using the Plausibility and Gain Values}
Given that a solution is plausible, the gain value allows us to quantify its utility for increasing the classification performance. Therefore, this measure is used for comparing solutions from different cluster levels, after discarding the ones that are not plausible enough. That is, we choose the plausible solution with the highest gain to find the best number of clusters. Furthermore, the gain measure may also help to quantify the utility of different sensors or features in case when multiple are available, which would help
selecting the ones that can best contribute to the problem.
%\newcomment{(not BS) ADD SOMETHING AFTER TALKING TO DAVID}

The parameter $B$ denotes the minimal number of data instances allowed in a leaf of the cluster-based decision tree that is built by our method in order to find labels for clusters. Decreasing this parameter means allowing a more fine-grained search for single-cluster regions which are needed to reveal labels of clusters that are overlapped by others. While decreasing $B$ often leads to more single-cluster regions, it also increases the sensitivity to noise.

\subsection{Evaluation on Real Sensor Data}

\newcommand{\numsubjects}{13}
\newcommand{\numcombos}{768}

\newcommand{\plotwidth}{0.45\textwidth}
\newcommand{\imagedir}{4_label_inferring}
\newcommand{\imagetype}{sweepErr}
\newcommand{\hist}[3]{
	\subfloat[#1]{
%	  \framebox{
%	    \parbox{\plotwidth}{
	      \includegraphics[trim= 5mm 15mm 5mm 18mm,clip=true,width=\plotwidth,type=pdf,ext=.pdf,read=.pdf]{#2}
%	      \includegraphics[width=\plotwidth,type=pdf,ext=.pdf,read=.pdf]{\imagedir/spider_#2_#3_\imagetype}
%	    }
%	  }
	\label{fig:results_hist_#3}
	}
}

\begin{figure*}[htbp]
\begin{center}
%\hist{label inferring, $B=10$}{4_label_inferring/hist_all_simple_B_10_sweepErr_noMean}{simple_B10}
\hist{label inferring, $B=5$}{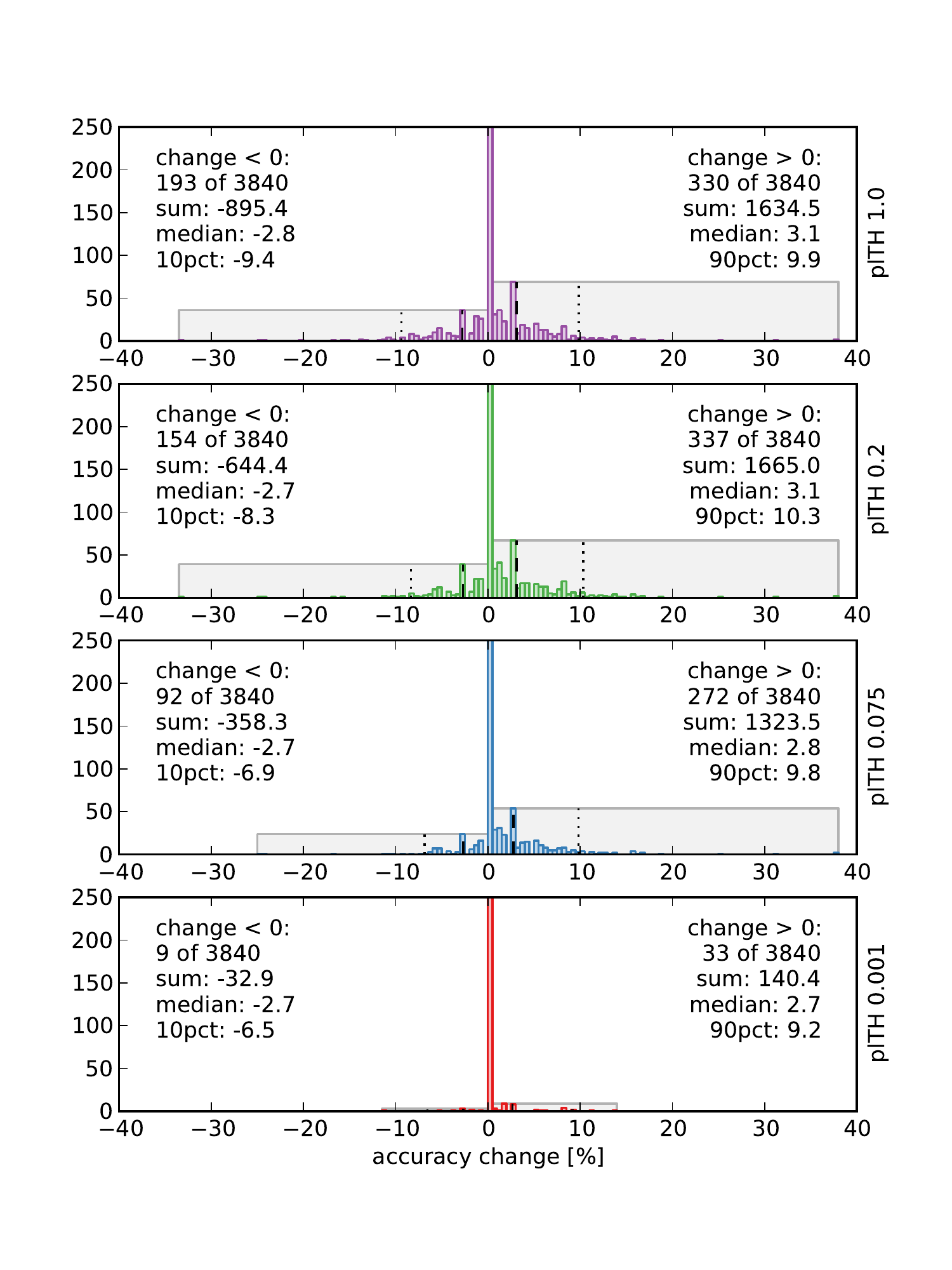}{simple_B5}
\hist{label inferring, $B=2$}{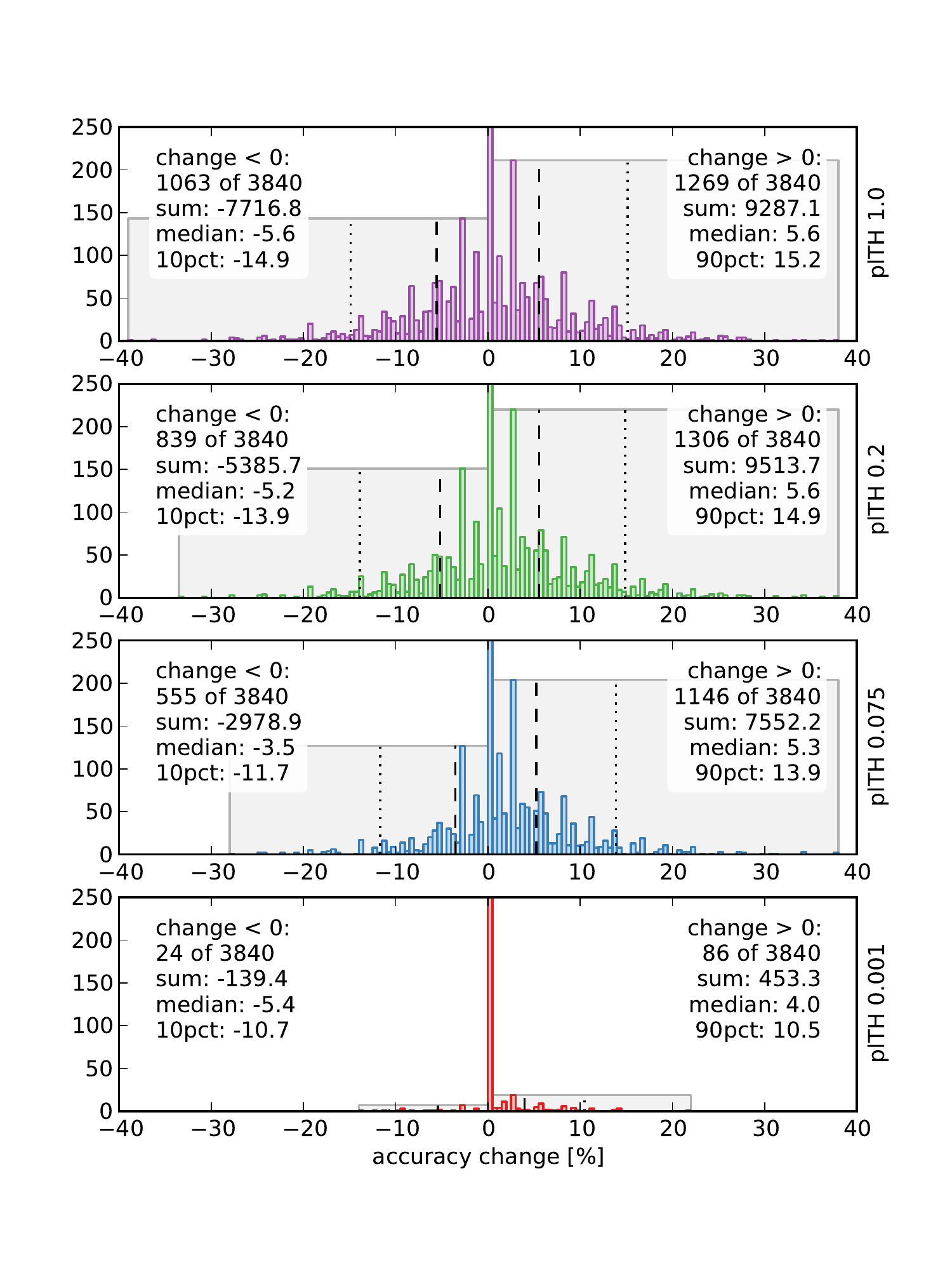}{simple_B2}
\\
\hist{similarity search, $B=10$}{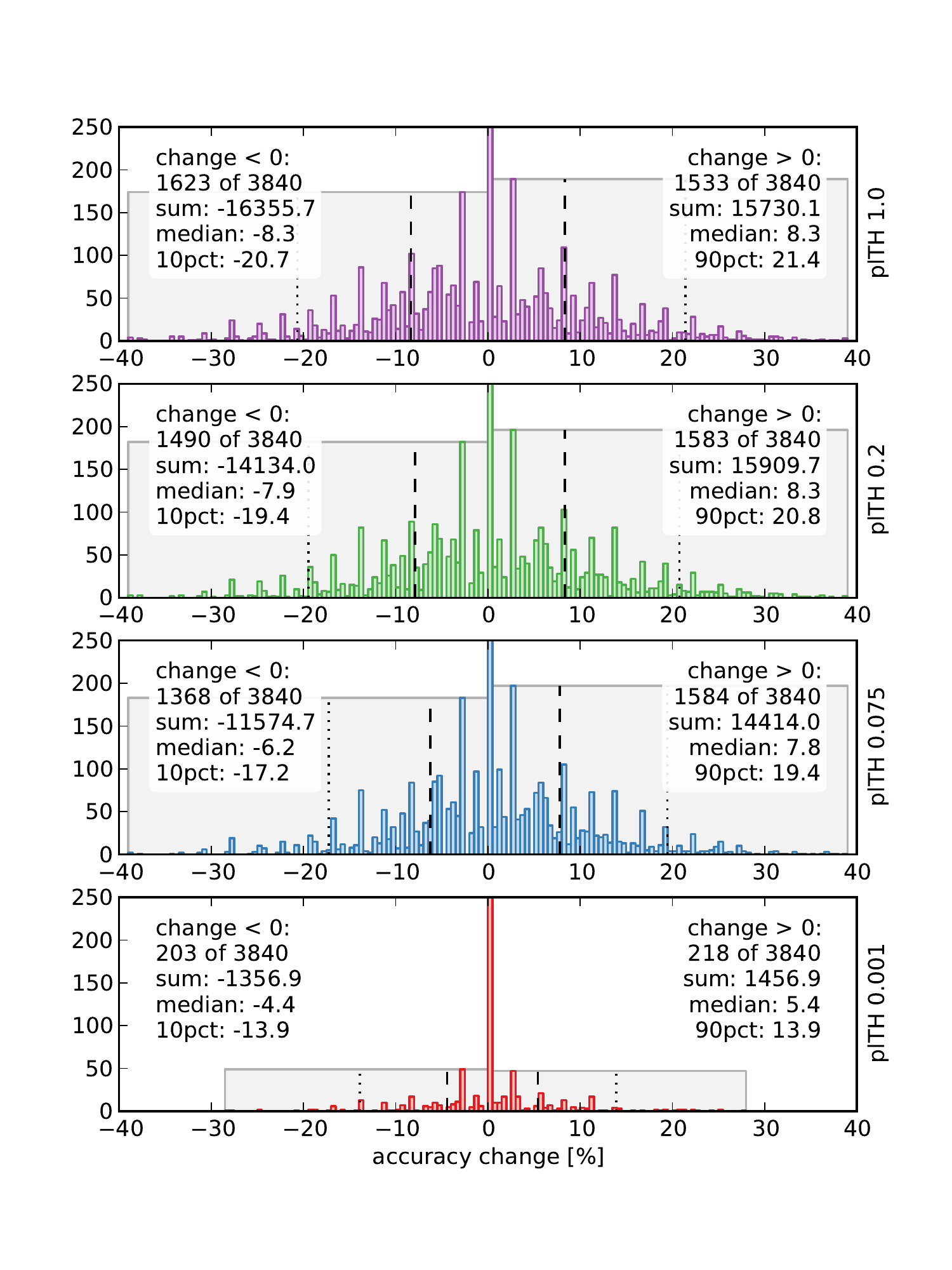}{both_B10}
\hist{similarity search, $B=2$}{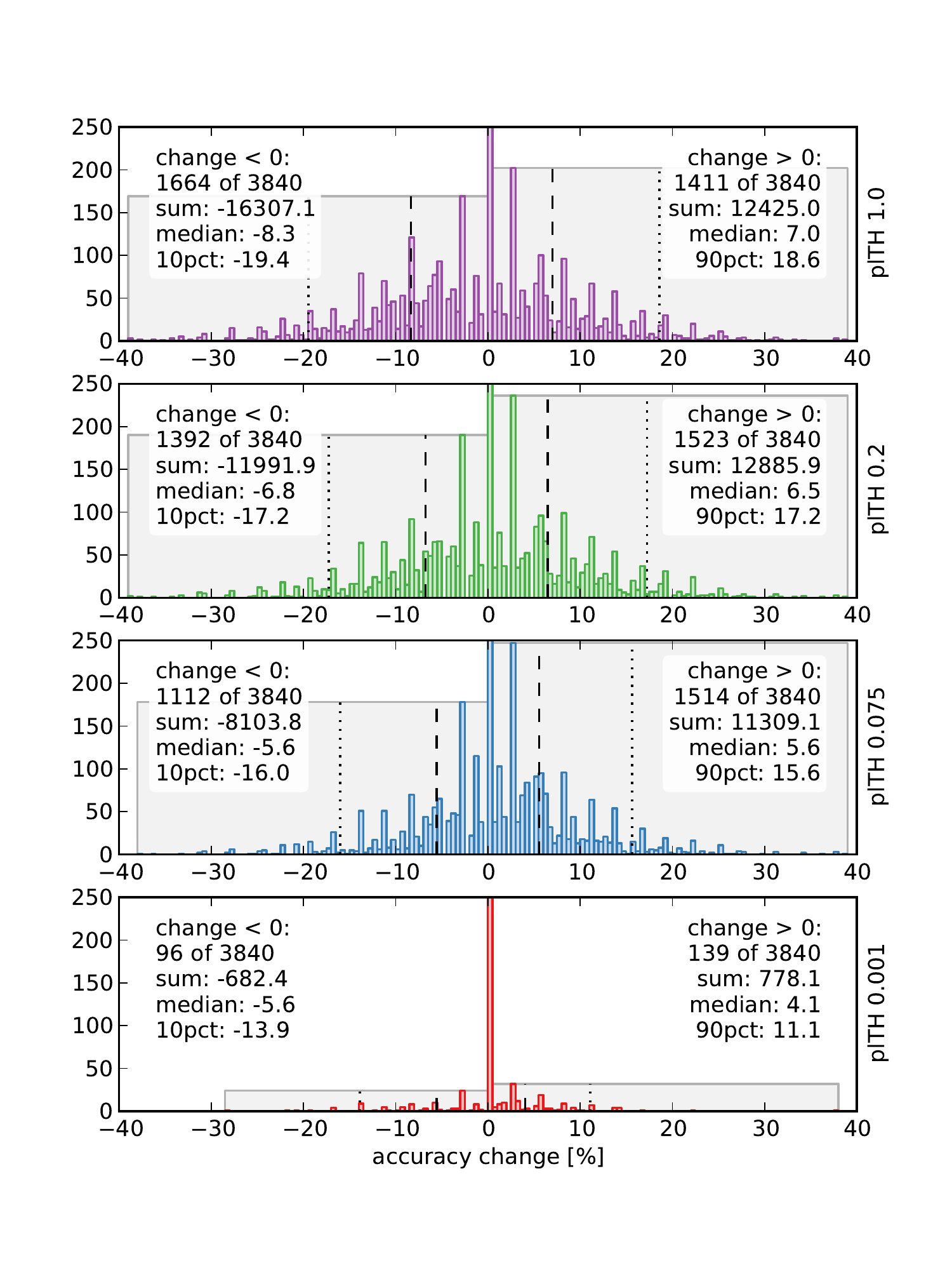}{both_B2}
\caption{Results for all variants with different parameterizations 
  on the three real-world data sets (13 subjects, 768*5 sensor combinations in total).
  Each plot shows results for four plausibility thresholds (plTH) 
  or four levels of user provided labels respectively.
  Bounding boxes for positive and negative results are highlighted, incl.\ 50/90 percentiles.}
%\caption{(continued)}
%\label{fig:results_realdata}
\end{center}
\end{figure*}

\begin{figure*}[htbp]
\ContinuedFloat
\begin{center}
\hist{bagging 0.9}{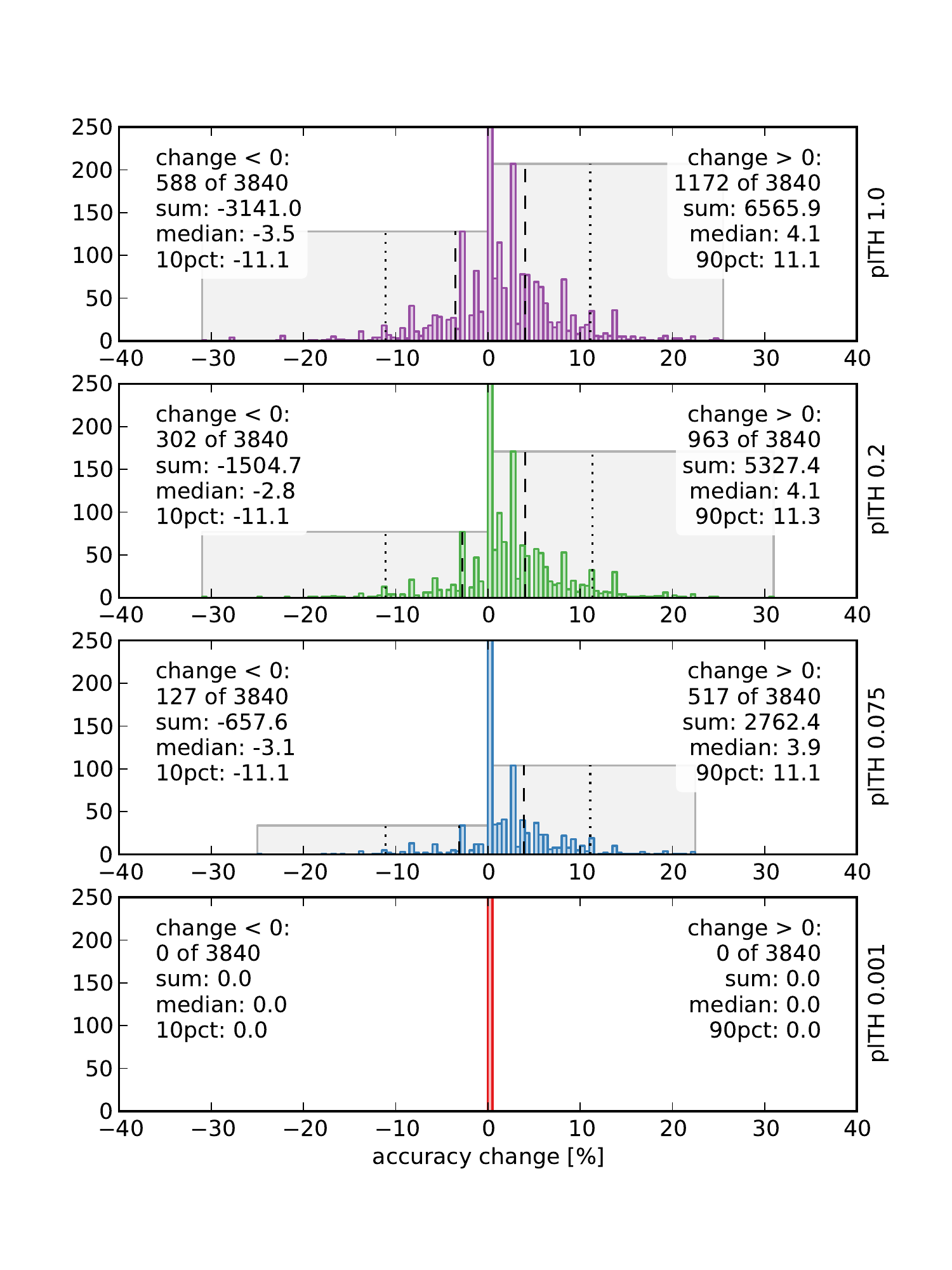}{bagging_09}
\hist{bagging 0.7}{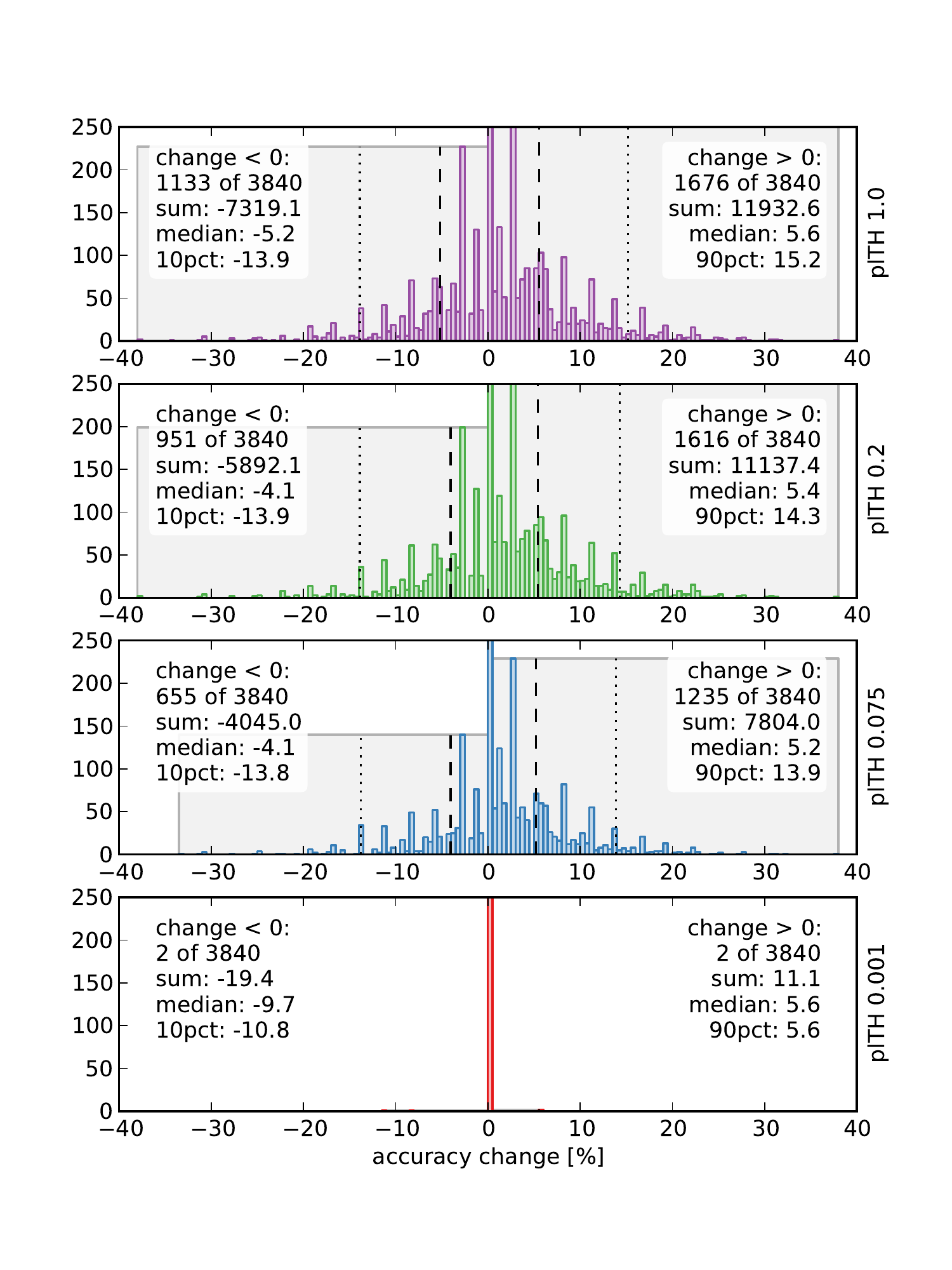}{bagging_07}
\\
\hist{pre-labeling, {\footnotesize plTH 0.2}}{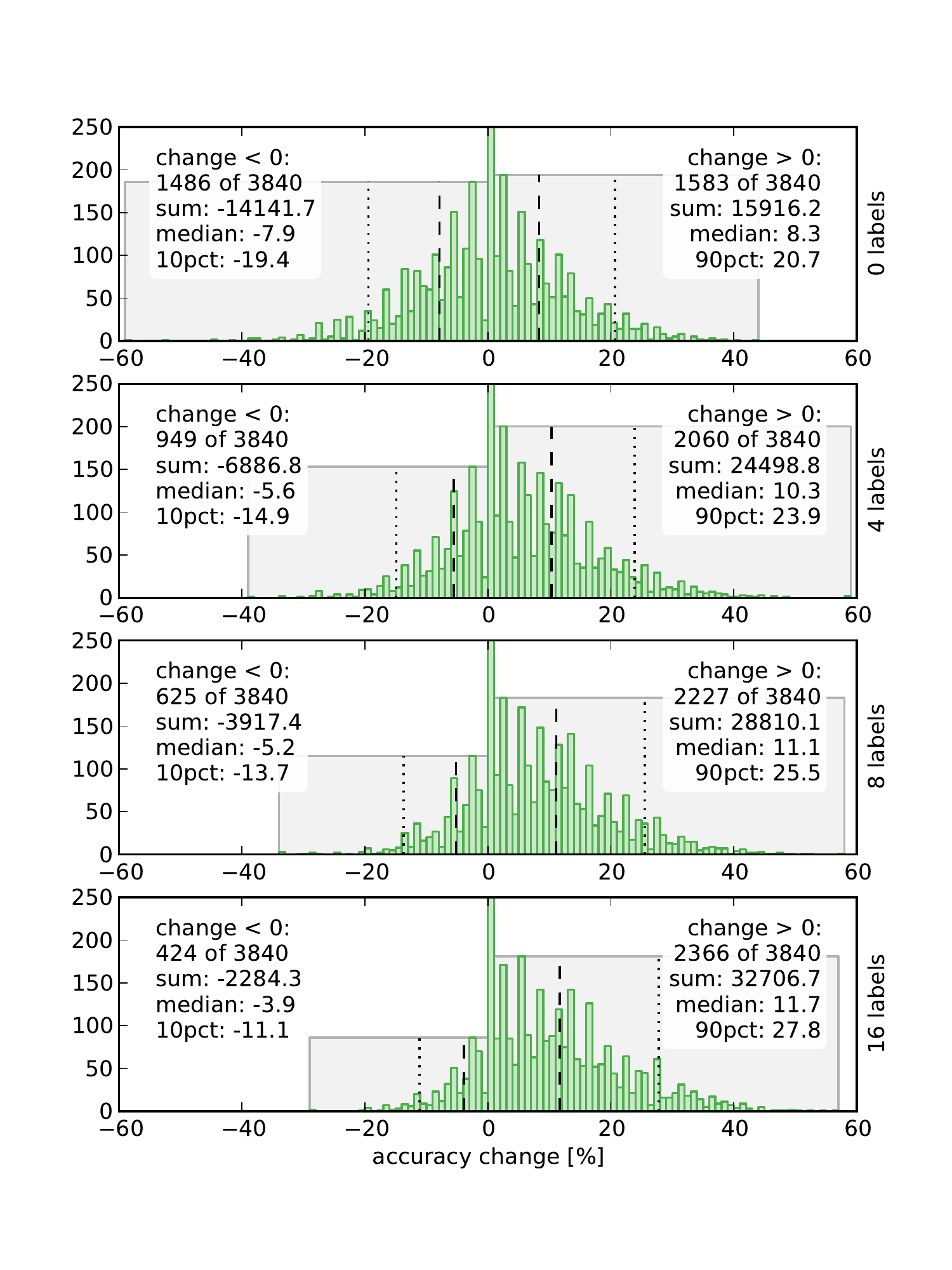}{prelabeling}
\hist{fault reduction, {\footnotesize plTH 0.075}}{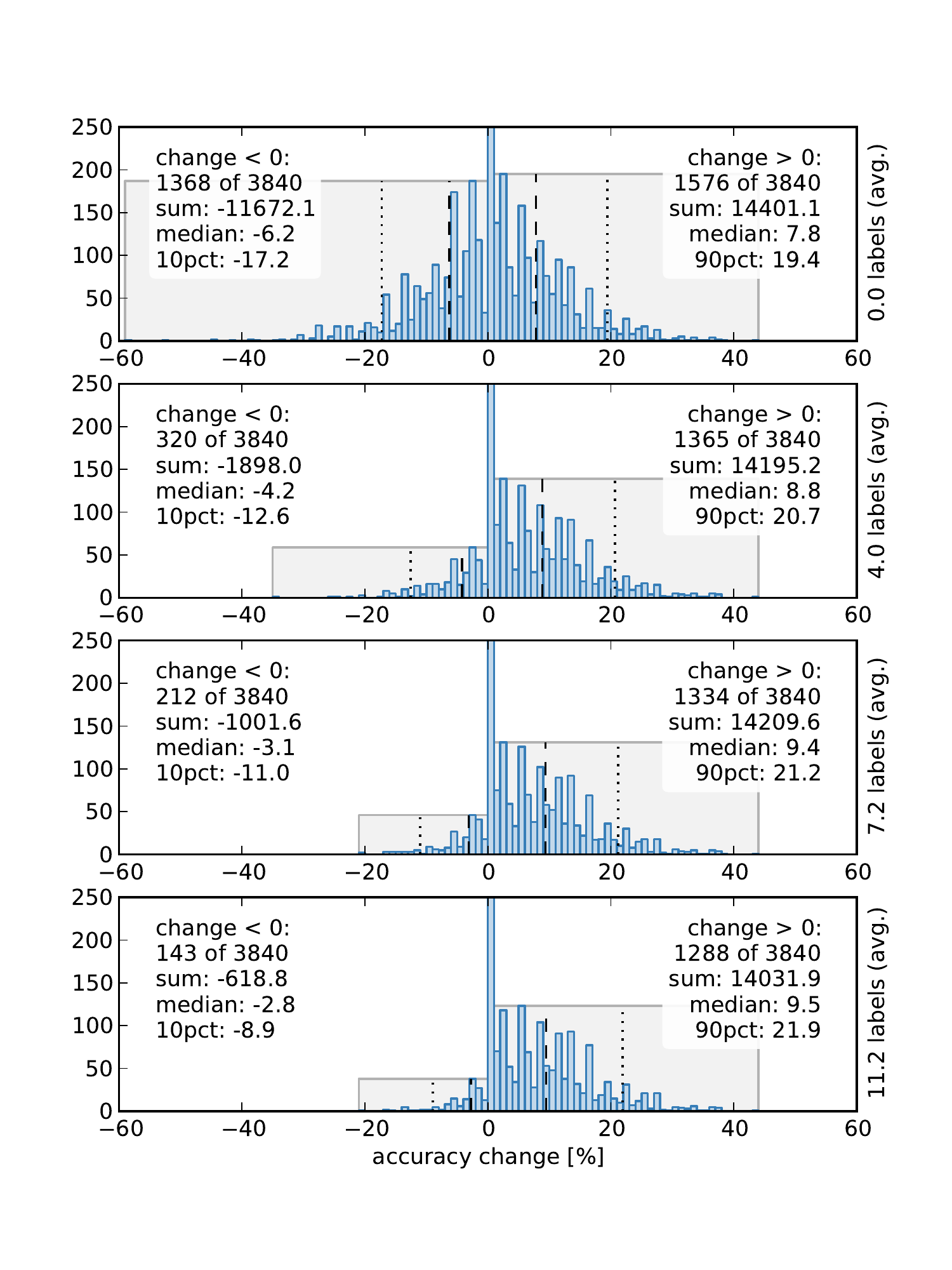}{fault_reduction}
\caption{(continued)}
%\label{fig:results_realdata}
\end{center}
\end{figure*}

\begin{figure*}[htb]
\ContinuedFloat
\begin{center}
\hist{pre-lab. + fault red., {\footnotesize plTH 0.2}}{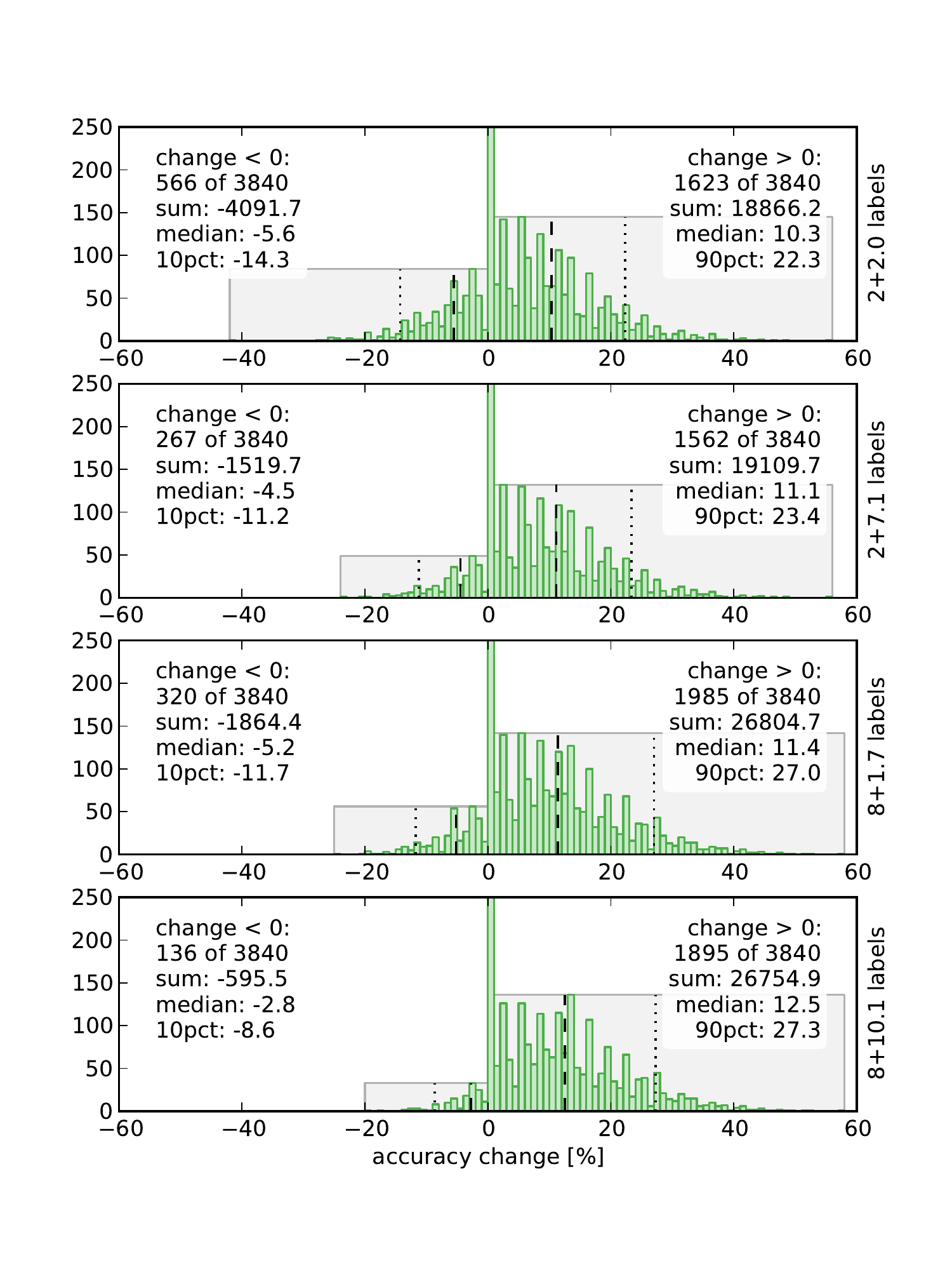}{pre+fault_reduction}
%\caption{Results for all variants with different parameterizations 
%  on the three real-world data sets (13 subjects, 768*5 sensor combinations in total).
%  Each plot shows results for four plausibility thresholds (plTH) 
%  or four levels of user provided labels respectively.
%  Bounding boxes for positive and negative results are highlighted, incl.\ 50/90 percentiles.}
\caption{(continued)}
\label{fig:results_realdata}
\end{center}
\end{figure*}

The results on all three real data sets are summarized in Fig.~\ref{fig:results_realdata}. The histograms display the number of sensor combinations per accuracy change, which is the unsupervised change of accuracy achieved by our method. They display the number of combinations with desirable (positive accuracy change) and undesirable (negative accuracy change) results. Regarding Fig.~\ref{fig:results_hist_simple_B2}, lowering the purity threshold (plTH) from $1.0$ (unrestricted, top) down to $0.001$ (strongly restricted, bottom) is clearly reducing the number and grade of undesirable results, while keeping nearly all desirable ones. Only at the lowest threshold ($0.001$), the number of desirable results is strongly reduced too.
This effect is also visible in the results with less fine grained cluster-based trees (Fig.\ \ref{fig:results_hist_simple_B5}), but rising the parameter $B$, i.e., increasing the minimal allowed leaf size for the cluster-based tree used for finding cluster labels, reduces not only undesirable results but also many of the desirable ones.

Judging from the histograms, the ``blue'' result in Fig.\ \ref{fig:results_hist_simple_B2} (plTH 0.075, $B$=2) seems the most reasonable. While it features many high grade desirable results (1146 of 3840) with a median of 5.3\%, it only has 555 (14.5\%) undesirable ones with a smaller median of -3.5\% accuracy change.
%
%The effects of both parameters (plTH, $B$) are almost identical when compared
%between separate results from all three data sets, suggesting that they will occur on other data sets correspondingly without need for adapting the parameters.

\section{Extension 1: Similarity Search}
%\section{Labeling Clusters Based On Distribution Similarity}
\label{sec:similarity_search}

In Section \ref{sec:distribution} and Fig.\ \ref{fig:clusters}, we have discussed the problem of ``hidden clusters''. These are clusters that have no points that are projected onto a single cluster region of the original feature space and, thus, can not be assigned a class distribution using the method described in the previous section.
%The effect of such hidden clusters can be seen in the performance graph in Fig.\ \ref{fig:dist10_100_perf}.
%Our method needs the cluster centers to be at least $\var{dist}{=}1.6$ apart  to closely approach the accuracy of supervised~2D training and achieve a nearly perfect improvement.
%Below a value of $\var{dist}{=}1$ there is nearly no improvement compared to the original classifier. 

\subsection{The Similarity Search Method}
\label{sec:similarity_search_method}
In this section, we describe an extension of the system that addresses the hidden cluster problem, step 2c in Fig.\ \ref{fig:flowchart} (bottom). The core idea is to perform an exhaustive search of all possible labellings of all hidden clusters with the aim of maximizing the plausibility (as defined in  Section \ref{sec:plausibility}) of the overall configuration. The search is done on the level of {\em multi-cluster regions} of the original feature space (as defined in Section \ref{sec:multicluster}). 

As illustration of the principle, consider a multi-cluster region of the original feature space where {\em all but one} of the clusters that are projected onto it from the new feature space already have class distributions assigned to them (see Fig.\ \ref{fig:similaritysearch}). All we need to do for estimating the label distribution of the unlabeled cluster is:
\begin{enumerate}
\item Retrieve the class distribution that the respective multi cluster region of the old feature space had in the original training data.
\item Compute the class distribution that results from the projection of the data from all the {\em labeled} clusters onto the respective multi-cluster region.
\item Compute the difference between the two above distributions. This difference is the distribution in the unlabeled cluster.
\end{enumerate} 
The above is based on the fact that plausibility is defined to be maximal when the distribution from the original training data is identical with the distribution resulting from the projection of points from the respective clusters onto the old feature space. 

Of course, the above method does not work when the set of clusters that are projected onto a multi-cluster region (call it $R_{1}$) contains more than one unlabeled cluster (consider two  unlabeled clusters $A$ and $B$).  In this case, the difference between the distributions will correspond to the class distributions in the union of clusters $A$ and $B$ with no information about the distribution within the individual clusters. 

A possible solution stems from the fact that, in general, each cluster is projected onto more than one multi-cluster region. Thus, cluster $A$ may also be projected onto a region $R_{2}$. If within $R_{2}$ cluster $A$ is the only unlabeled cluster, then it can be assigned a distribution. This means that now, within $R_{1}$, $B$ is now the only unlabeled cluster and can also be assigned a distribution.

In the general case, we consider the {\em dependency graph} rather than just cluster pairs and individual regions. A dependency graph has a node for each cluster and edges connecting each pair of clusters that appear together in the same region. We iteratively traverse the dependency graph starting with nodes that we can label and then propagating the information along the edges to label further nodes and so on. The method terminates when propagating the information from a newly labeled cluster does not lead to any  nodes being assigned to a distribution. Note that termination does not necessarily mean that unambiguous distributions were found for all clusters. However, as the results presented in the next section will show, a significant number of clusters can be labeled and the method clearly outperforms the naive approach described in the previous section.

A final concern with respect to similarity search is the computational complexity. A complete search over all possible labellings leads to a combinatoric explosion: with $N$ classes and $K$ clusters there are $N^K$ candidates to verify for each region. However, we do not perform such a complete search in general. The connected components of the dependency graph form disjoint groups of clusters. The sets of regions in which the groups of clusters appear are disjoint, too. This means that the labeling of the clusters in one group does not affect the labeling of any other cluster and we may search the best labeling for each group of clusters individually. Thus, there are maximally $N^{\max(|g_i|)}$ candidates to verify, where $|g_i|$ denotes the number of clusters in the $i$-th group.
The maximal size of connected components $\max(|g_i|)$ is expected to be small in activity recognition scenarios. It is related to the maximal number of classes which the initial classifier confuses.

\subsection{Evaluation on Real Sensor Data}

For the evaluation on real world data sets, we proceed as described in Section \ref{sec:evaluation}. Figs.\ \ref{fig:results_hist_both_B10} and \ref{fig:results_hist_both_B2} provide a detailed overview of these results with different levels of purity thresholds applied. The results confirm the key observations from the previous section: the similarity search leads to higher improvement but also carries a higher risk of a performance decrease when faced with a data set that violates the basic assumptions on which our method is built. Thus, for example, with $B{=}2$ and $\var{plTH}{=}0.075$ (which is the optimal case) we can improve the resulting accuracy in more cases (1514 of 3840---39.4\%, as opposed to 1146 (29.8\%))
 with more  improvement
(8.5\% on average with the best 10\% thereof improving more than
15.6\% as opposed to 6.6\% and 13.9\%) than with the simple approach. However,  
on the negative side we also get more cases (1112 (29.0\%) as opposed
to 555) with more average
accuracy loss 
(median -5.6\% as opposed to -3.5\%).

\section{Extension 2: Bagging}
\label{sec:bagging}

The results of the similarity search method presented in the previous section reveal a potentially significant performance improvement, but also a sensitivity to noise in data sets. Here, we show how the well-known bagging~\cite{bagging} technique can be adapted to our approach to prevent our method from being misled by noise and sample variance.

\subsection{The Bagging Method}
\label{sec:bagging_method}
When a small change in the data set produces a different solution, i.e., because of a cluster getting labeled differently or the clustering deviates, this is a strong indication of either an ambiguous situation or over-fitting (high model complexity). Both are cases that we want to avoid. Thus, filtering out those solutions is desirable, and it is exactly what can be achieved with bagging.

Thus, before starting from the given unlabeled data set, we form $10$ replicas of the same size by drawing each instance at random, but with replacement. Each instance may appear repeated times or not at all in any of the replica. We run our method individually on each replicated data set and aggregate the resulting classifiers. Where at least a certain percentage of the classifiers agree (i.e., defined by a \emph{bagging threshold}), the majority result is taken. Otherwise, we fall back to the result from the initial classifier without any improvement.

In the case of decision tree classifiers on which we base our experiments, we can merge the aggregated classifiers resulting from the bagging step into a single decision tree. This simplifies classification afterwards, as only a single tree hast to be traversed.

Our method of integrating a new sensor splits up the regions of the original classifier, which are the leaves in the decision tree along the new dimension of the feature space. Thus, each leaf may get replaced by a subtree, providing new leaves for each split region. When merging the aggregated classifiers, we first overlay the new subtrees in each region and successively (1) assign the majority label to all split regions (leaves of the subtree) where at least \emph{bagging threshold} percent of the subtrees agree, and (2) assign the initial label of that region (from the original classifier) to all other split regions. Adjacent split regions with identical label are joined.

\subsection{Evaluation on Real Sensor Data}

The improvement obtained in real sensor data is presented in Figs.\ \ref{fig:results_hist_bagging_07} and \ref{fig:results_hist_bagging_09}. The more relevant results were obtained with a threshold of 0.9. For a threshold of 0.7, bagging with similarity search results in a range comparable to that observed for the simple method described in Section \ref{sec:label_inferring}; i.e., with less negative but also less positive cases (for example for plTH~0.075 and B=2 we have positive results in 1235 cases vs.\ 1148 for the simple method and 1514 for the similarity search, while negative results are for 655 cases vs.\ 555 for the simple method and 1112 for similarity search). On the other hand, for a threshold of 0.9, the negative results are reduced much more than the positive ones. Thus, the sum of positive results ($\var{change}{>}0$) is then nearly 4 times the sum of negative results (e.g., 5327.4 vs.~-1504.7 for plTH~0.2). Also, with plTH~0.075 there are just 3.3\% (127) of cases with a negative accuracy change and in 13.5\% (517) of the cases we achieve an improvement of 5.3\% on average. In summary, the combination of similarity search and bagging (with a threshold of 0.9) brings the system in line with the ``doing much more good than harm'' vision and makes it useful for real-life applications.

\section{Extension 3: Pre-Labeling}
\label{sec:prelabeling}

The challenge that we address in this section is how our method performs if the user is providing a small amount of reliable labels for the instances recorded with the additional sensor.

\subsection{The Pre-Labeling Method}

We extend the process depicted in Fig.\ \ref{fig:flowchart}. In particular, we add an extra step just before the cluster labels are determined by the label inferring and similarity search methods. Note that, unlike classical semi-supervised methods (e.g.~\cite{Nigam_EM}), we do not expect to have enough labeled instances to directly assign class distributions to all clusters. Instead, we assume that clusters labeled from user-provided information will function as ``anchors'', increasing the effectiveness of the similarity search. As described in Section~\ref{sec:similarity_search}, similarity search can assign a class distribution to an unlabeled cluster if it is the only unlabeled cluster in a multi-cluster region. At the same time, it considers dependency graphs that have clusters as nodes and multi cluster regions as edges. The information about the newly assigned distribution is propagated along such a graph, reducing the number of unlabeled clusters in other multi cluster regions. Thus, at best, labeling a single cluster can set off a cascade, leading to all clusters in a previously fully unlabeled dependency graph to be assigned correct distributions. Obviously, this will not be the default case. In other cases, there may not be a cascade effect at all. However, the assumption is that on average there will be a significant multiplicative effect so that a small number of user provided labels will lead to a significant performance improvement.  

\subsection{Evaluation on Real Sensor Data}

When applied to the real world data sets, the positive effect of pre-labeling is also clearly visible. Fig.\ \ref{fig:results_hist_prelabeling} shows the results of pre-labeling in combination with the similarity search method (no bagging, plTH 0.2). The number of cases with negative accuracy change decreases from 1486 (38.7\%) down to 949 (24.7\%) with just four labels and to 625 (16.3\%) with eight labels (1.15 labels per class on average). At the same time, the number of positive outcomes increase from 1583 (41.2\%) to 2060 (53.6\%) for four labels and to 2227 (58\%) for eight labels. The improvement is particularly well visible in the sum over all improvements which goes from 15916 with no labels to 24499 with four and 28810 with eight.

\renewcommand{\imagedir}{8_fault_reduction}

\section{Extension 4: Fault Reduction}
\label{sec:fault_reduction}
The pre-labeling method from the previous section is meant to increase the number of clusters that the similarity search method can label---which essentially means increasing the amount of improvement that the system can generate. The reduction in the number of sensor combinations where it caused a performance decrease was a ``by-product'', as additional correctly labeled clusters offset wrong labels in the statistics. By contrast, in this section, we present a method for using user-provided labels to directly detect and discard classifier variants that lead to performance degradation.

\subsection{The Fault Reduction Method}

The general idea is based on the observation that to assess whether a classifier using an additional sensor increases or decreases classification accuracy, we only need to consider data points  where the new and the old classifier disagree. As a consequence we 
\begin{enumerate}
\item Run the ``old'' (1D) and the ``new'' (2D: old plus new sensor) classifier ($C_{1}, C_{2}$) in parallel.
\item Actively ask the user for labels for instances where the classifiers disagree.
\item Reject the new classifier if it fails a {\bf test criterion} defined as being correct at least a certain predefined number of times.  
 In general, to keep the number of labels small, we consider only a few instances of disagreement between the old and the new classifier and require the new one to be correct in at least 75\% of the cases.  
\end{enumerate}
The fault reduction method is applied right after all cluster configurations have been labeled and assigned final gain and plausibility values (step 3 in Fig.\ \ref{fig:flowchart} (top)). Without fault reduction, a solution with the best gain/plausibility trade-off (according to application-specific criteria) would be chosen (if it exists) and all other solutions discarded. With fault reduction, we take a few (we found three to work well) best solutions and run them in parallel with the old classifier as described above. If a solution fails the test criterion, it is discarded. If in the end all solutions have been discarded, we consider the sensor combination to be not useful and do not use it (we get no improvement but also run no risk of a performance decrease). Otherwise, we pick the one with the best gain/plausibility trade-off from the ``surviving'' classifiers.

\subsection{Success Probability Estimation}
Alike all the methods presented in this article, fault reduction is a heuristic that works often but can not be guaranteed to be always right. Thus, we may have a new classifier that, overall, leads to a significant performance degradation, but the instances that we acquire labels for (which are random) happen to be just the ones on which it is right. On the other hand, we may reject a very good new classifier because of picking just the very few instances on which it is wrong. However, as shown below, the strength of our method is its theoretic model that is significantly ``skewed'' in favor of keeping good classifiers and rejecting poor ones given a random sample of sequences.

\begin{figure}[htb]
\begin{center}
\includegraphics[width=0.85\columnwidth, type=pdf,ext=.pdf,read=.pdf]{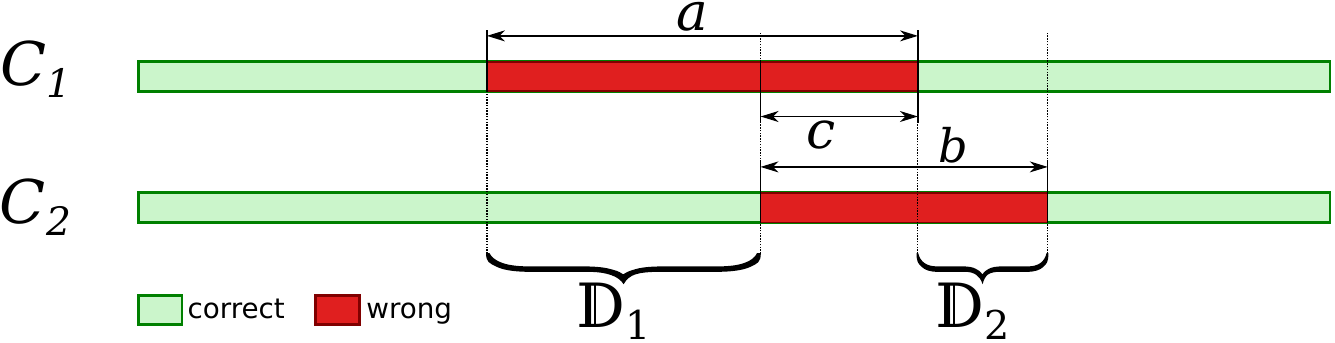}
\end{center}
\caption{Schematic of two classifiers $C_1$ and $C_2$ applied to test data with areas of disagreement.}
\label{fig:compare-classifiers}
\end{figure}

More formally, given two classifiers that disagree exactly at the instances where $C_{1}$ is wrong but $C_{2}$ is correct ($\mathds{D}_{1}$), and where $C_{2}$ is wrong but $C_{1}$ is correct ($\mathds{D}_{2}$), we face a situation as illustrated by Fig.\ \ref{fig:compare-classifiers}.

The probability that $C_{2}$ is correct for a randomly drawn instance from
$\mathds{D} = \mathds{D}_{1} \cup \mathds{D}_{2}$
hence is
\begin{equation}
P_{C_{2}}
\! = \! P(x \notin \mathds{B} \mid x \in \mathds{D})
\! = \! \frac{\left| \mathds{D}_{1} \right|}{\left| \mathds{D}_{1} \right| + \left| \mathds{D}_{2} \right|}
\! = \! \frac{a-c}{(a-c)+(b-c)},
\end{equation}
where $\mathds{B}$ is the set of all instances classified wrongly by $C_{2}$,
$a$ the number of instances classified wrong by $C_{1}$,
$b = \left| \mathds{B} \right|$, and
$c$ the number of instances classified wrongly by both classifiers ($0 < c < \min(a,b)$).
When expressing the improvement as $v = a-b$, we receive
\begin{equation}
P_{C_{2}} = \frac{a-c}{2(a-c)-v}.
\end{equation}

If we accept the classifier $C_{2}$ only in case it is correct on all $n$ randomly drawn instances from $\mathds{D}$, we can express the probability of accepting the classifier as a function of the improvement $v$:
\begin{equation}
p(v)
% = P_{C_{2}}^n
 = \left(\frac{a-c}{2(a-c)-v}\right)^n.
\end{equation}
From this function, we see that better performing classifiers ($v > 0$) are more likely to be accepted than those that perform worse than the initial classifier ($v < 0$). This effect is amplified with increasing number of tested instances $n$, but at the same time classifiers with small improvement tend to be rejected as well. With larger ``overlap'' $c$, the bad classifiers ($v<0$) are strongly filtered out and even classifiers with small improvement get higher probability of being accepted, yet the maximal possible improvement ($v \leq a-c$) is more restricted by the ``overlap''. With two identical classifiers ($c{=}a,v{=}0$) the function is undefined. In that case, there are no disagreement instances to test with ($\mathds{D}{=}\emptyset$).

\subsection{Evaluation on Real Sensor Data}

For evaluation purposes, we select the user-provided labels randomly (cf.\ selection strategies for active learning). This is achieved by just selecting the $n$ first instances of the adaptation set that meet the method's requirements and augmenting them with ground truth labels (the dataset is shuffled before splitting into the three parts for training, adaptation, and testing). As before, the evaluation is performed five times on each sensor combination with differently shuffled data sets.

As measure of the amount of user provided information, we use the number of disagreements between the old and the new classifiers that have to be considered for the test criterion. Note that since we are running the test on several of the new classifiers this is not necessarily equal to the number of labels. In the best case this may be equal, but in the worst case we need a different set of labels of each of the new classifiers. Thus, in the evaluation we provide the average number of labels that the system used, not the setting for the number of disagreements (which is our control parameter).

The histograms in Fig.\ \ref{fig:results_hist_fault_reduction} confirm this effect for the real-world data sets. An average of four labels already reduces the number of faults from 1368 (35.6\%) to just 320 (8.3\%) while still keeping 1365 (35.5\%) occasions where the accuracy is improved. With 11.2 user-provided labels the faults further decrease to 3.7\% without affecting the positive outcomes too much (33.5\%). Moreover, when combining pre-labeling (with eight labels) and fault reduction with 10.1 labels (Fig.\ \ref{fig:results_hist_pre+fault_reduction}), the positive outcomes can even be increased to 49.3\% with just 3.5\% of negative ones.

\section{Conclusion and Outlook}
\label{sec:conclusion}

In this article, we presented and evaluated novel techniques that enable activity recognition systems to self-adapt to new sensor information becoming available in their environment. While these techniques show promising results on real data (and on artificial data, see \cite{bannach2015tools} for more details), we are aware that their practical applicability is still quite limited. First, only an extension from one sensor to a second was investigated. In practice, multi-sensor systems are used and the transition from a low to a high dimensional space is anything but trivial. Second, the approach so far consists of a limited number of heuristics applied to a specific classifier type, with no theoretical underpinning and comparison to other methods. Finally, many important aspects such as feature selection, online adaptation, long term evaluation, and overall system architecture were not investigated. 
%That is, much further fundamental research is needed.

Having recognized this, we defined a roadmap for our future research addressing the following key challenges: (1) How can appropriate features for the AR tasks be chosen autonomously? (2) Assuming that generative, probabilistic classifiers with static system behavior shall be used for AR: How can self-adaption be realized? (3) Assuming that discriminative, dynamic classifiers shall be used for AR: How can the solution from above be extended? (4) How can humans be integrated at run-time in a very efficient and effective way by means of active learning techniques? (5) How can we ensure that continuous modifications of the system configuration lead to long term  improvement and not  to unbounded performance degradation of the overall system?  (6) Which quality metrics are needed and how can the various new techniques be evaluated in AR case studies? 
Altogether, our research will provide new technologies for self-improvement (if an additional sensor can be used) and self-healing (if a sensor that dropped out can be replaced by another) of multi-modal sensor systems in the field of AR.

\section*{Acknowledgment}
This work was supported by the EU-funded OPPORTUNITY project (FET-Open grant 225938).
The authors would like to thank Dr.\ Tobias Reitmaier for providing information on active and semi-supervised learning.
The work has been funded by the German Research Foundation (DFG) within the project ``Organic Computing Techniques for Run-Time Self-Adaptation of Multi-Modal Activity Recognition Systems'' (LU 1574/2-1 and SI 674/12-1).



%% References


\bibliographystyle{elsarticle-num}
\bibliography{newsensor}

\begin{thebibliography}{10}
\expandafter\ifx\csname url\endcsname\relax
  \def\url#1{\texttt{#1}}\fi
\expandafter\ifx\csname urlprefix\endcsname\relax\def\urlprefix{URL }\fi
\expandafter\ifx\csname href\endcsname\relax
  \def\href#1#2{#2} \def\path#1{#1}\fi

\bibitem{onbody}
K.~Kunze, P.~Lukowicz, H.~Junker, G.~Tr{\"o}ster, Where am {I}: Recognizing
  on-body positions of wearable sensors, in: LNCS, Vol. 3479, 2005, pp.
  264--275.

\bibitem{orientation}
K.~Kunze, P.~Lukowicz, K.~Partridge, B.~Begole, Which way am i facing:
  Inferring horizontal device orientation from an accelerometer signal, in:
  Proc. of Int. Symp. on Wearable Computers (ISWC), {IEEE} Press, 2009, pp.
  149--150.

\bibitem{kaiposition}
K.~Kunze, P.~Lukowicz, Dealing with sensor displacement in motion-based onbody
  activity recognition systems, in: Proceedings of the 10th International
  Conference on Ubiquitous Computing, 2008, pp. 20--29.

\bibitem{magnetic}
K.~Kunze, G.~Bahle, P.~Lukowicz, K.~Partridge, Can magnetic field sensors
  replace gyroscopes in wearable sensing applications?, in: Proc. 2010 Int.
  Symp. on Wearable Computers, 2010, pp. 1--4.

\bibitem{jaenicke14self}
M.~J\"anicke, B.~Sick, P.~Lukowicz, D.~Bannach, Self-adapting multi-sensor
  systems: A concept for self-improvement and self-healing techniques, in: SASO
  Workshops, 2014, pp. 128--136.

\bibitem{Pan:2010dm}
S.~J. Pan, Q.~Yang, {A Survey on Transfer Learning}, Knowledge and Data
  Engineering, IEEE Transactions on 22~(10) (2010) 1345--1359.

\bibitem{jiang2008literature}
J.~Jiang, A literature survey on domain adaptation of statistical classifiers,
  URL: http://sifaka. cs. uiuc. edu/jiang4/domainadaptation/survey.

\bibitem{daume2009frust}
H.~D. III, \href{http://arxiv.org/abs/0907.1815}{Frustratingly easy domain
  adaptation}, CoRR abs/0907.1815.
\newline\urlprefix\url{http://arxiv.org/abs/0907.1815}

\bibitem{DaumeIII:2007wo}
H.~Daum{\'e}, III, Frustratingly easy domain adaptation, ACL.

\bibitem{Duan:2012wg}
L.~Duan, D.~Xu, I.~Tsang, {Learning with Augmented Features for Heterogeneous
  Domain Adaptation}, arXiv.org\href {http://arxiv.org/abs/1206.4660v1}
  {\path{arXiv:1206.4660v1}}.

\bibitem{minnen2007discovering}
D.~Minnen, T.~Starner, M.~Essa, C.~Isbell, {Discovering characteristic actions
  from on-body sensor data}, in: Proc. IEEE ISWC 2006, 2007, pp. 11--18.

\bibitem{huynh2006unsupervised}
T.~Huynh, B.~Schiele, {Unsupervised discovery of structure in activity data
  using multiple eigenspaces}, Springer LNCS (2006) 151--167.

\bibitem{wyatt2005unsupervised}
D.~Wyatt, M.~Philipose, T.~Choudhury, Unsupervised activity recognition using
  automatically mined common sense, in: Proceedings of the 20th National
  Conference on Artificial Intelligence - Volume 1, AAAI'05, AAAI Press, 2005,
  pp. 21--27.

\bibitem{tapia2006building}
E.~Tapia, T.~Choudhury, M.~Philipose, {Building reliable activity models using
  hierarchical shrinkage and mined ontology}, Pervasive Computing (2006)
  17--32.

\bibitem{chen2009ontology}
L.~Chen, C.~Nugent, {Ontology-based activity recognition in intelligent
  pervasive environments}, International Journal of Web Information Systems
  5~(4) (2009) 410--430.

\bibitem{stikic2008exploring}
M.~Stikic, K.~Van~Laerhoven, B.~Schiele, {Exploring semi-supervised and active
  learning for activity recognition}, in: IEEE ISWC 2008., 2008, pp. 81--88.

\bibitem{guan2007activity}
D.~Guan, W.~Yuan, Y.~Lee, A.~Gavrilov, S.~Lee, {Activity Recognition Based on
  Semi-supervised Learning}, in: Proc. of the 13th IEEE Int. Conference on
  Embedded and Real-Time Computing Systems and Applications, 2007, pp.
  469--475.

\bibitem{CalmaRSL15}
A.~Calma, T.~Reitmaier, B.~Sick, P.~Lukowicz, A new vision of collaborative
  active learning, CoRR abs/1504.00284.

\bibitem{RV95}
J.~Ratsaby, S.~S. Venkatesh, Learning from a mixture of labeled and unlabeled
  examples with parametric side information, in: Proceedings of the eighth
  annual conference on Computational learning theory, COLT '95, ACM, New York,
  NY, USA, 1995, pp. 412--417.

\bibitem{chapelle2006semi}
O.~Chapelle, B.~Scholkopf, A.~Zien, {Semi-supervised learning}, Vol.~2, MIT
  Press, 2006.

\bibitem{BD98}
K.~Bennet, A.~Demiriz, Semi-supervised support vector machines, in: Advances in
  Neural Information Processing Systems 11, MIT Press, 1998, pp. 368--374.

\bibitem{GCMC08}
L.~Gomez-Chova, G.~Camps-Valls, J.~Munoz-Mari, J.~Calpe, Semisupervised image
  classification with laplacian support vector machines, Geoscience and Remote
  Sensing Letters, IEEE 5~(3) (2008) 336 --340.
\newblock \href {http://dx.doi.org/10.1109/LGRS.2008.916070}
  {\path{doi:10.1109/LGRS.2008.916070}}.

\bibitem{zhu2006semi}
X.~Zhu, {Semi-supervised learning literature survey}, Computer Science,
  University of Wisconsin-Madison.

\bibitem{Angluin88}
D.~Angluin, Queries and concept learning, Machine Learning 2~(4) (1988)
  319--342.

\bibitem{ACLEM90}
L.~Atlas, D.~Cohn, R.~Ladner, M.~A. El-Sharkawi, R.~J. Marks, II, Training
  connectionist networks with queries and selective sampling, in: Advances in
  Neural Information Processing Systems 2, Morgan Kaufmann, Denver, CO, 1990,
  pp. 566--573.

\bibitem{LG94}
D.~D. Lewis, W.~A. Gale, A sequential algorithm for training text classifiers,
  in: Proceedings of the Seventeenth Annual International ACM SIGIR Conference
  on Research and Development in Information Retrieval (SIGIR '94), Dublin,
  1994, pp. 3--12.

\bibitem{Settles09}
B.~Settles, Active learning literature survey, Computer {S}ciences {T}echnical
  {R}eport 1648, University of Wisconsin, Department of Computer Science
  (2009).

\bibitem{TK02}
S.~Tong, D.~Koller, Support vector machine active learning with applications to
  text classification, Journal of Machine Learning Research 2 (2002) 45--66.

\bibitem{CL06}
C.~Constantinopoulos, A.~Likas, Active learning with the probabilistic {RBF}
  classifier, in: Artificial Neural Networks--ICANN 2006, Vol. 4131 of Lectures
  Notes in Computer Science, Springer, Berlin, Germany, 2006, pp. 357--366.

\bibitem{SC08}
B.~Settles, M.~Craven, An analysis of active learning strategies for sequence
  labeling tasks, in: Proceedings of the Conference on Empirical Methods in
  Natural Language Processing (EMNLP '08), Honolulu, HI, 2008, pp. 1070--1079.

\bibitem{NS04}
H.~T. Nguyen, A.~Smeulders, Active learning using pre-clustering, in:
  Proceedings of the Twenty-first International Conference on Machine Learning,
  ICML '04, ACM, New York, NY, USA, 2004, pp. 79--86.
\newblock \href {http://dx.doi.org/10.1145/1015330.1015349}
  {\path{doi:10.1145/1015330.1015349}}.

\bibitem{DCB07}
P.~Donmez, J.~G. Carbonell, P.~N. Bennett, Dual strategy active learning, in:
  Proceedings of the 18th European Conference on Machine Learning (ECML '07),
  Warsaw, Poland, 2007, pp. 116--127.

\bibitem{MMP04}
P.~Mitra, C.~A. Murthy, S.~K. Pal, A probabilistic active support vector
  learning algorithm, IEEE Transactions on Pattern Analysis and Machine
  Intelligence 26~(3) (2004) 413--418.

\bibitem{GG07}
Y.~Guo, R.~Greiner, Optimistic active learning using mutual information, in:
  Proceedings of the 20th International Joint Conference on Artifical
  Intelligence (IJCAI '07), Hyderabad, India, 2007, pp. 823--829.

\bibitem{DRH06}
C.~K. Dagli, S.~Rajaram, T.~S. Huang, Utilizing information theoretic diversity
  for {SVM} active learning, in: Proceedings of the 18th International
  Conference on Pattern Recognition (ICPR '06), Hong Kong, China, 2006, pp.
  506--511.

\bibitem{Dai:2007tu}
W.~Dai, G.-R. Xue, Q.~Yang, Y.~Yu, {Co-clustering based classification for
  out-of-domain documents}, in: Proceedings of the 13th ACM SIGKDD
  international conference on Knowledge discovery and data mining, ACM, 2007,
  pp. 210--219.

\bibitem{Blum:1998ua}
A.~Blum, T.~Mitchell, {Combining labeled and unlabeled data with co-training},
  in: Proceedings of the eleventh annual conference on Computational learning
  theory, ACM, 1998, pp. 92--100.

\bibitem{calatroni2010methodology}
A.~Calatroni, D.~Roggen, G.~Troster, {A methodology to use unknown new sensors
  for activity recognition by leveraging sporadic interactions with primitive
  sensors and behavioral assumptions}, in: Proc. of the Opportunistic
  Ubiquitous Systems Workshop,UBICOMP 2010, 2010.

\bibitem{THSRSWS14}
S.~Tomforde, J.~H{\"{a}}hner, H.~Seebach, W.~Reif, B.~Sick, A.~Wacker,
  I.~Scholtes,
  \href{http://ieeexplore.ieee.org/xpl/articleDetails.jsp?arnumber=6775093}{{Engineering
  and Mastering Interwoven Systems}}, in: {ARCS} 2014 - 27th International
  Conference on Architecture of Computing Systems, Workshop Proceedings,
  February 25-28, 2014, Luebeck, Germany, University of Luebeck, Institute of
  Computer Engineering, 2014, pp. 1--8.
\newline\urlprefix\url{http://ieeexplore.ieee.org/xpl/articleDetails.jsp?arnumber=6775093}

\bibitem{THS14}
S.~Tomforde, J.~H{\"a}hner, B.~Sick,
  \href{http://dx.doi.org/10.1007/s00287-014-0827-z}{{Interwoven Systems}},
  Informatik-Spektrum 37~(5) (2014) 483--487, {Aktuelles Schlagwort}.
\newblock \href {http://dx.doi.org/10.1007/s00287-014-0827-z}
  {\path{doi:10.1007/s00287-014-0827-z}}.
\newline\urlprefix\url{http://dx.doi.org/10.1007/s00287-014-0827-z}

\bibitem{MSSU11}
C.~M\"uller-Schloer, H.~Schmeck, T.~Ungerer (Eds.), Organic Computing -- A
  Paradigm Shift for Complex Systems, Birkh\"auser Verlag, 2011.

\bibitem{bagging}
L.~Breiman, Bagging predictors, Machine Learning 24~(2) (1996) 123--140.
\newblock \href {http://dx.doi.org/10.1007/BF00058655}
  {\path{doi:10.1007/BF00058655}}.

\bibitem{bicycle1}
G.~Ogris, T.~Stiefmeier, H.~Junker, P.~Lukowicz, G.~Troster, {Using Ultrasonic
  Hand Tracking to Augment Motion Analysis Based Recognition of Manipulative
  Gestures}, in: Proc. IEEE ISWC 2005, 2005, pp. 152--159.

\bibitem{skoda}
T.~Stiefmeier, D.~Roggen, G.~Ogris, P.~Lukowicz, G.~Tr{\"o}ster, Wearable
  activity tracking in car manufacturing, IEEE Pervasive Computing 7~(2) (2008)
  42--50.

\bibitem{opportunity}
D.~Roggen, K.~Forster, A.~Calatroni, T.~Holleczek, Y.~Fang, G.~Troster,
  P.~Lukowicz, G.~Pirkl, D.~Bannach, K.~Kunze, et~al., {OPPORTUNITY: Towards
  opportunistic activity and context recognition systems}, in: Proc. WoWMoM
  2009., IEEE, 2009, pp. 1--6.

\bibitem{Nigam_EM}
K.~Nigam, A.~K. Mccallum, S.~Thrun, T.~Mitchell, Text classification from
  labeled and unlabeled documents using em, Machine Learning 39 (2000)
  103--134.
\newblock \href {http://dx.doi.org/10.1023/A:1007692713085}
  {\path{doi:10.1023/A:1007692713085}}.

\bibitem{bannach2015tools}
D.~Bannach, Tools and methods to support opportunistic human activity
  recognition, Ph.D. thesis, Technical University of Kaiserslautern (2015).

\end{thebibliography}






% \end{thebibliography}

\end{document}